\documentclass[Afour,sageh,times]{sagej}

\usepackage{moreverb,url}
\usepackage{multicol}
\usepackage{caption,tabularx,booktabs}
\usepackage[utf8]{inputenc} 
\usepackage[T1]{fontenc}    
\usepackage{hyperref}       
\usepackage{url}            
\usepackage{booktabs}       
\usepackage{amsfonts}       
\usepackage{nicefrac}       
\usepackage{microtype}      
\usepackage{xcolor}         
\usepackage{times}
\usepackage{multirow}
\usepackage{graphicx}
\usepackage{color}
\usepackage{array}
\usepackage{algorithm}
\usepackage{algorithmic}
\usepackage{amssymb}
\usepackage{amsmath}
\usepackage{stfloats}
\usepackage{wrapfig}
\usepackage[justification=centering]{caption}

\usepackage[normalem]{ulem}
\newcolumntype{P}[1]{>{\centering\arraybackslash}p{#1}}
\usepackage{float}
\usepackage{natbib}
\usepackage[font=small,skip=8pt]{caption}

\newcommand\BibTeX{{\rmfamily B\kern-.05em \textsc{i\kern-.025em b}\kern-.08em
T\kern-.1667em\lower.7ex\hbox{E}\kern-.125emX}}

\begin{document}
\runninghead{}

\title{Semantically Controllable Augmentations for Generalizable Robot Learning}

\author{Zoey Chen*\affilnum{1}, Zhao Mandi*\affilnum{2}, Homanga Bharadhwaj*\affilnum{3,4}, \\ Mohit Sharma\affilnum{3}, Shuran Song\textsuperscript{\dag}\affilnum{2}, Abhishek Gupta\textsuperscript{\dag}\affilnum{1}, Vikash Kumar\textsuperscript{\dag}\affilnum{3}}

\affiliation{\affilnum{1}University of Washington, USA\\
\affilnum{2}Stanford University, USA \\
\affilnum{3}Carnegie Mellon University, USA \\
\affilnum{4}FAIR, AI at Meta \\
$^*$ equal contribution \\
$^\dagger$ equal advising \\}

\corrauth{Zoey Chen, Homanga Bharadhwaj}

\email{qiuyuc@cs.washington.edu, hbharadh@cs.cmu.edu}

\begin{abstract}
Generalization to unseen real-world scenarios for robot manipulation requires exposure to diverse datasets during training. However, collecting large real-world datasets is intractable due to high operational costs. For robot learning to generalize despite these challenges, it is essential to leverage sources of data or priors beyond the robot's direct experience. In this work, we posit that image-text generative models, which are pre-trained on large corpora of web-scraped data, can serve as such a data source. These generative models encompass a broad range of real-world scenarios beyond a robot's direct experience and can synthesize novel synthetic experiences that expose robotic agents to additional world priors aiding real-world generalization at no extra cost. 

In particular, our approach leverages pre-trained generative models as an effective tool for data augmentation. We propose a generative augmentation framework for semantically controllable augmentations and rapidly multiplying robot datasets while inducing rich variations that enable real-world generalization. Based on diverse augmentations of robot data, we show how scalable robot manipulation policies can be trained and deployed both in simulation and in unseen real-world environments such as kitchens and table-tops. By demonstrating the effectiveness of image-text generative models in diverse real-world robotic applications, our generative augmentation framework provides a scalable and efficient path for boosting generalization in robot learning at no extra human cost.
\end{abstract}

\maketitle

\section{Introduction}

While robot learning has often focused on the search for plausible policies~\citep{levine15endtoend, nagabandi:corl2019} or motions plans ~\citep{qureshi18mpn} in specific scenarios, the benefits of learning methods in robotics come from the prospect for \emph{generalization}. Going beyond policy optimization in highly controlled situations such as warehouses or factories, robot learning methods have the potential for widespread generalization across tasks, environments, and objects. While techniques such as imitation learning methods circumvent the challenges of exploration, teaching a robot various skills requires a large amount of experience and diverse data sources. Different from collecting vision and language data, robot demonstration data requires active interaction with the scene. This is thus expensive, and prior works have indeed spent years collecting large robot manipulation datasets for imitation, through techniques like tele-operation~\cite{rt1}. Beyond the total quantity of data, the rigidity of most robotics setups makes it non-trivial to collect \emph{diverse} data in a wide variety of scenarios. As a result, many robotics datasets involve a single setup with just a few hours of robot data. 

The limitation of data diversity has been a challenge for the field of machine learning in general. Training reliably effective models primarily hinges on access to datasets that comprehensively represent the target environment. Beyond the challenges of scale, informational diversity while pivotal is hard to capture. These limitations often impede the model's ability to generalize effectively to unseen scenarios. To get around these challenges, data augmentation techniques, such as color adjustments, Gaussian blur, and cropping, have traditionally been exploited to enhance the generalization capabilities of machine learning models. 

These techniques have also proven effective in the field of robot learning for handling minor variations in appearances (color, lighting, etc). They however fall short in addressing structured variations in the scene such as the introduction of distractors, alterations in the background, or changes in the object's visual appearance. These limitations arise from their inability to provide augmentations introducing diverse, realistic, and semantic alterations in the data, which are crucial for training robust policies capable of adapting to diverse unseen real-world scenarios. These considerations are particularly important in the field of robotics, where the availability of data is often constrained due to operational and safety challenges. The ability to simulate a wide range of realistic and semantically meaningful scenarios is therefore crucial for generalization.

\begin{figure*}[!h]
    \centering
    \includegraphics[width=\textwidth]{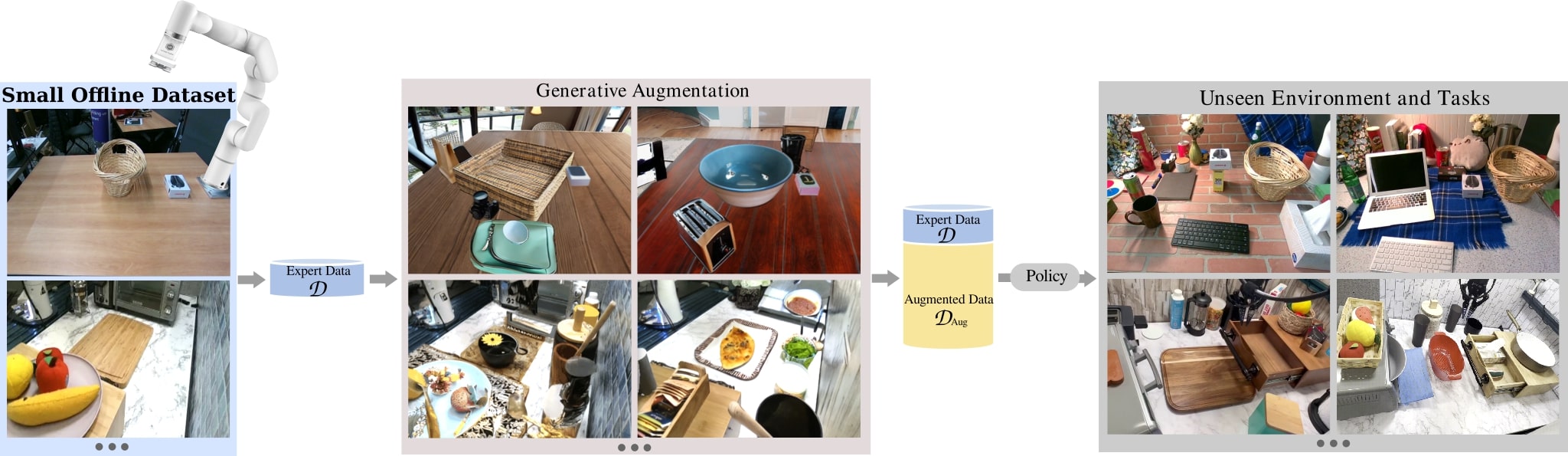}
    \caption{Our Framework takes a small offline dataset containing expert demonstrations and leverages text-to-image generative models to semantically bootstrap the initial dataset into a much larger and diverse augmented dataset, which can be used to train a robot policy that generalizes to unseen environments and tasks.}
    \label{fig:main_figure}
    \vspace{-1.5em}
\end{figure*}
In this work, we introduce a framework for \emph{semantic} data augmentation, which aims at automatically and significantly enabling broad robot generalization, by leveraging pre-trained generative models. While on-robot data can be limited, the data that pre-trained generative models are exposed to is significantly larger and more diverse ~\citep{schuhmann22laion, deng09imagenet} such as LAION5B~\citep{laion5b} dataset. Our work aims to leverage these generative models as a source of data augmentation for real-world robot learning, and expose robots to a broader spectrum of experiences than through direct data collection itself. This method crucially imposes invariances in the model against a range of semantic variations, effectively equipping robots with the adaptability required for real-world applications.

The limited on-robot experience offers crucial demonstrations of the target behavior, but the true depth of our approach lies in how a generative model enriches this learning. By creating a wide array of visual scenes featuring varied backgrounds and object appearances, the generative models actively enforce invariances in the learning agent. This ensures that the desired behavior remains consistent and valid across these diverse and semantically rich environments.

This allows us to cheaply generate a large quantity of \emph{semantically augmented} data from a small number of demonstrations, providing a learning agent access to significantly more diverse scenes than the purely on-robot demonstration data. As we show empirically, this can lead to widely improved generalization, with minimal additional burden on human data collection.

Given a dataset of image-action examples provided on a real robot system, we automatically augment the original robot observation for entirely different and realistic environments, which display the visual realism and complexity of scenes that a robot might encounter in the real world. In particular, our framework leverages language prompts with a generative model to change object textures and shapes, adding new distractors and background scenes in a way that is physically consistent with the original scene. These augmented data together with the corresponding new language descriptions are used for training robots and generalize to unseen environments.  

We show that training on this \emph{semantically augmented} dataset significantly improves the generalization capabilities of imitation learning methods in entirely unseen real-world environments. We train language-conditioned policies in both single-task and multi-task table-top settings and show in-depth experiments and discussions in both real-world robot experiments and simulation. Our experiments analyze the generalization of the trained policies at different levels and demonstrate the overall benefits of generative augmentations in robot manipulation across tasks, and settings. 

\section{Related Work}

\paragraph{\textbf{Variance Injection into Learning}} 

The concept of injecting invariance into learning models has been employed in prior works. Domain randomization, for instance, exposes physics invariances but relies on access to parametric models of the environment. Our work focuses on visual generalization—a domain where access to such environmental parameters is often not feasible. 
The most widely used technique for injecting visual variance is various forms of data augmentation ~\citep{shorten19aug}, such as cropping, shifting, noise injection and rotation. These methods have been used in many robot learning approaches and provide a significant improvement in data efficiency \citep{benton2020learning, cubuk2018autoaugment, shorten2019survey, kostrikov2020image}. For example, \citep{zafar2022comparison} investigate different augmentation modes in Meta-learning settings. In addition, several methods attempt to enforce geometric invariance through architectural innovations such as ~\citep{wang22equivariant} and \citep{deng2021vector}. 
While these methods can provide a local notion of robustness and invariance to perceptual noise, they do not provide generalization to novel object shapes or scenes. 
More recently out-of-domain models have started making their way into robot learning. For example - \citep{kapelyukh2022dall} uses large text-image models like Dall-e \citep{ramesh2022hierarchical} to generate favorable image goals for robots.

These approaches while helpful in task specification, provide limited benefit for robots to generalize to entirely unseen situations. In contrast, our framework induces semantic changes to the observations thereby helping acquire behavior invariance to new scenes.

\paragraph{\textbf{Alternate Data Sources in Robotics.}}
The recent advancements in self-supervised methods across language and vision fields have shown the benefits of utilizing extensive datasets. A lot of recent works have studied the use of pre-trained visual representations trained mainly on datasets of non-robot interactions~\citep{ego4d,imagenet}, for learning control policies \citep{r3m, moco, shridhar2022cliport,majumdar2023we, shah2021rrl}. Many works focus on single-task settings \citep{r3m, moco, sharma2023lossless, hansen2022pre}, or simulated robot environments~\citep{hansen2022pre,majumdar2023we}. 
Given challenges with collecting \emph{large} real-world robotics datasets, some works focus on alternate data sources such as language~\citep{tellex2011understanding, lynch2020language, stepputtis2020language, saycan}, human videos \citep{nguyen2018translating, bharadhwaj2023towards, zhou2021manipulator,shao2021concept2robot,shaw2023videodex,bharadhwaj2023zero,bahl2022human,bahl2023affordances,track2act,wang2023mimicplay}, goal image generations~\cite{bharadhwajvisual,susie,dallebot}, and generative augmentations \citep{rao2020rl, kapelyukh2022dall, rosie}.

\paragraph{\textbf{Visual Policy Learning }} 

In the realm of robot learning, the choice of data modality is critical for achieving generalization. Vision data became particularly important because it captures intricate details necessary for complex tasks such as spatial reasoning and object manipulation. visual data can effectively form the backbone of effective robotic control policies, as shown by recent works~\citep{visual1,visual2,visual3,dreamer,visual_imitation1,visual_imitation2,visual_imitation3,r3m,shridhar2022cliport}. One key step in training a generalizable visual policy for robots is to collect and train on diverse data such that the policies are robust and adaptable to understanding and interacting with their environment effectively.
Recent studies have explored ways to expand the volume and variety of visual data for robot learning. A significant portion of this research is centered on gathering and analyzing data directly generated by robots~\citep{rt1,rt2}. However, these works typically involve only a limited number of distinct environments, posing challenges for the robots to generalize effectively across a broader spectrum of unfamiliar scenes. A substantial body of research also focuses on learning image representations beyond robot demonstration data such as large-scale videos and images~\citep{r3m,vip,mvp}. Moreover, several studies have focused on using language to learn representations from videos \citep{zhao2022lavila, momeni2023verbs}. 

Unlike the constrained settings of direct robot-generated data or specific external datasets, we argue the importance of visual diversity is not only in volume but also in semantic richness. To this end, we leverage pre-trained generative models, that can synthetically generate a wide array of complex and varied visual scenes. Our framework actively embeds invariance into the original data and is key to enabling robots to generalize across a much broader spectrum of scenes, including those they have never directly experienced.

\paragraph{\textbf{Scaling Robot Learning }}  Recent advancements in robot learning have utilized self-supervised learning \citep{agpinto, lynch2020learning, berscheid2019improving} and simulations \citep{metaworld, james2020rlbench, mittal2023orbit, zhu2020robosuite} to craft versatile multi-purpose agents. These developments span both simulated \citep{gato, vima, schrittwieser2020mastering, espeholt2018impala, sodhani2021multi, kaiser2019simple} and real environments \citep{tobin2017domain,shridhar2022cliport,handa2023dextreme,robocat}. However, a gap remains: multi-task RL in narrow simulated domains \citep{espeholt2018impala, song2019v} and limited real-world generalization \citep{mtrf}. While some initiatives \citep{gato, vima,gradientsurgery} explore diverse scenarios, their focus remains largely on simulation-based policy evaluation. 
Learning a multi-task agent requires extensive scaling of data diversity to ensure broad generalization. This is where the potency of semantic augmentations becomes crucial, as they allow for the efficient cultivation of agents that are generalizable across a wide array of tasks. Our framework enables scaling from single-task learning with minimal demonstrations to complex multi-task environments that may encompass up to 7.5k demonstrations, effectively addressing the challenge of achieving robust multi-task learning without the need for prohibitively large datasets.

We present a unified framework based on~\citep{genaug,cacti,bharadhwaj2023roboagent} showing the benefits of generative augmentations across different spectrums of structure and scale. At one extreme, we show how we can inject invariance into learning with low data as few as a single demonstration per task, and develop generalizable single task policies. At the other extreme, we show how we can scale automatic generative augmentations that preserve less structure but can be applied to large-scale datasets for learning generalizable language-conditioned multi-task policies that work reliably across diverse scenes.

\section{Background and Formulation}

Here, we describe the problem statement we consider in our semantic data augmentation technique - Generative Augmentation and show how generative models can conceptually be used to inject semantic invariances into the robot learning frameworks. Shown in Figure \ref{fig:main_figure}, we aim to bootstrap an initial small offline dataset using generative augmentation, and train a robot policy that generalizes widely on unseen environments and tasks. In this section, we will first formulate the problem of learning from demonstrations, followed by the proposed method of leveraging generative models for data augmentation.
\subsection{Problem Formulation}
Our work considers general robotic decision-making problems and we specifically focus on robot manipulation.
Our setup considers a robot arm that receives sensory observations $o \in \mathcal{O}$ such as camera observations, and outputs appropriate action $a \in \mathcal{A}$ (e.g. where to move the robot arm for picking up an object). Our goal is to learn a model (a policy) $f_\theta: \mathcal{O} \rightarrow \Delta{A}$ (where $\Delta{A}$ denotes the simplex over actions) that predicts a distribution over actions such that the action $a \sim f_\theta(.|o)$ can accomplish a task when executed in the environment. In this work, we restrict our consideration to supervised learning methods for learning $f_\theta(.|o)$. We assume a human expert provides a dataset of demonstrations $\mathcal{D} = \{(o_0, a_0), (o_1, a_1), \dots, (o_N, a_N)\}$ for solving different tasks. We use maximum likelihood training to learn optimal policies for the provided demonstrations \citep{zeng2020transporter, shridhar2021cliport}:

\begin{equation}
    \max_\theta \mathbb{E}_{(o, a) \sim \mathcal{D}}\left[ \log f_\theta(a|o) \right]
\end{equation}

As noted above, our training process is limited to the demonstration dataset $\mathcal{D}$ collected by the human supervisor. Since collecting large-scale human demonstration data is hard, the dataset size $|D|$ is most often quite limited. Data augmentation techniques are often used to increase the dataset size. Data augmentation methods apply augmentation functions $q: \mathcal{O} \times \mathcal{A} \times \mathcal{Z} \rightarrow \mathcal{O} \times \mathcal{A}$ which generate augmented data $(o', a') = q(o, a, z); z \sim p(z)$, where different noise vectors $z$ generate different augmentations. This could include augmentations like Gaussian noise, cropping, and color jitter amongst others \citep{benton2020learning, cubuk2018autoaugment, shorten2019survey, perez2017effectiveness}. Using this augmentation function we can sample a large number of different augmentations to create an augmented dataset $\mathcal{D}_{\text{aug}} = \mathcal{D} \cup \{(o', a')_i\}_{i=1}^M$, where $M \gg N$, and then used for maximum likelihood training of $f_\theta(a|o)$.
Typically, most augmentation functions $q$ are manually specified by researchers. 
Further, these functions don't add any new semantic meaning to data, but instead help prevent overfitting by making models robust to disturbances like color changes, shifts and rotations.
In the next section, we'll explore how generative models can be used for semantic data augmentation, creating more visually diverse and realistic data that better reflects the complexity of the real world.

\subsection{Leveraging Generative Models for Data Augmentation}
While data augmentation methods typically hand-define augmentation functions $(o', a') = q(o, a, z); z \sim p(z)$, the generated data $(o', a')$ may not be particularly relevant to the true distribution of real-world data. Since most of these generated variations do not appear in the real-world distribution it is unclear if generating such a large augmented dataset $\mathcal{D}_{\text{aug}}$ helps learned predictors $f$ generalize in real-world settings. By contrast, the key insight in our framework is that pre-trained text-to-image generative models such as Stable Diffusion \citep{stablediffusion}

are trained on the distribution $p_{\text{real}}(o)$ of real images (including real scenes that a robot might find itself in). This lends them the ability to generate (or modify) the training set observations $o$ in a way that corresponds to the distribution of real-world scenes instead of a heuristic approach such as described in \citep{perez2017effectiveness}. We will use this ability to perform targeted data augmentation for improved generalization of the learned predictor $f_\theta$. 

We formalize our augmentation setting by assuming access to generative models $g: \mathcal{T} \times \mathcal{O} \times \mathcal{Z} \rightarrow \mathcal{O}$, which map from an image $o$, a text description $t$, and a noise vector $z$ to a modified image $o' = g(o, t, z); z \sim p(z)$. This includes commonly used text-to-image inpainting models such as Make-A-Video \citep{singer2022make}, DALL-E 2 \citep{ramesh2022hierarchical}, Stable Diffusion \citep{rombach2022high} and Imagen \citep{saharia2022photorealistic}. 

While these generative models excel at creating novel visual observations $o$, they do not inherently generate new actions $a$. Instead, their strength lies in the potential to enforce \emph{semantic invariance} in the learned model $f_\theta$, ensuring that varied but semantically related observations ${o, g(t_1, o, z_1), g(t_2, o, z_2), \dots, g(t_M, o, z_M)}$ correspond to the same action $a$. To harness this potential in pre-trained text-to-image generative models for semantic data augmentation, we can generate sets of semantically equivalent observation-action pairs ${(o, a), (g(t_1, o, z_1), a), \dots}$ for each $(o,a) \in \mathcal{D}$, ensuring the generated observations maintain semantic equivalence with the original action $a$.

This enables generating a diverse dataset of \emph{semantically meaningful} augmentations while still performing the specific task in the respective trajectories. Unlike typical data augmentation with the hand-defined shifts described above, the generated augmented observations $\{g(t_1, o, z_1), g(t_2, o, z_2), \dots, g(t_M, o, z_M)\}$ have a high likelihood under the distribution of real images $p_{\text{real}}(o)$ that a robot may encounter on deployment. This ensures that the model generalizes to a wide variety of novel scenes, making it significantly more practical to deploy in real-world scenarios since it will be robust to changes in objects, distractors, backgrounds, and other characteristics of an environment. 
Although our approach allows us to create a large set of relevant augmentations it still has few limitations. First, our augmentations can only create new observations for provided actions and cannot generate novel actions $a$, Second, generating new observations without care can also sometimes lead to physically inaccurate augmentations, e.g., inaccurate contacts leading to collisions between objects, physical inconsistencies such as objects in air. In the next section we use a common table-top robotic manipulation to discuss our method in more detail.

\section{Generative Augmentations for Robot Learning}
 \label{sec:method}
We describe our framework for generative augmentations in robot learning, that enables training policies through behavior cloning for generalization to environments and tasks beyond the original demonstrations.  

\subsection{Framework Overview}

\textbf{Semantic Augmentation} -- In the initial phase, the pre-collected dataset is expanded by generating a variety of semantic augmentation to the robot's existing experiences. This process transforms a single or limited robotic demonstration into multiple versions, each containing different semantic elements like objects, textures, and backgrounds, without requiring additional human demonstrations. This approach of enriching data with real-world semantic variances enhances the multi-task agent's generalization to unforeseen, out-of-distribution scenes that the robot might encounter during test time.

\textbf{Policy Learning} --  
The second phase focuses on learning robust robot skills using a small amount of robot data. This is achieved by adapting design choices from previous works that were usually limited to single-task environments and applying them to achieve larger-scale adaptability across various multi-task and multi-scene manipulation tasks. In addition to the single-task policies, we introduce a multi-task, language-driven policy framework. These are designed to train versatile agents capable of acquiring a range of skills from diverse, multi-modal datasets.

\subsection{Semantic Data Augmentation}
We consider two different regimes for our semantic augmentation. The first is a low-data regime that is controllable over the structure of semantic augmentations, including the use of 3D meshes and segmentation masks. This gives more control over augmentations and allows more physically plausible augmentations. However, its manual aspects limit its scalability and make it less feasible to augment extensive datasets. The second is a regime with large data, with a framework for completely automatic augmentations using pre-trained models. This approach enables us to scale up to datasets comprising thousands of trajectories, facilitating the learning of extensive multi-task policies, which can be deployed in diverse scenarios based on a specified goal. We detail each regime in the following sections.

\subsubsection{Structure-Aware Augmentation for Low-Data Regime}
\begin{figure*}[!h]
    \centering
    \includegraphics[width=\textwidth]{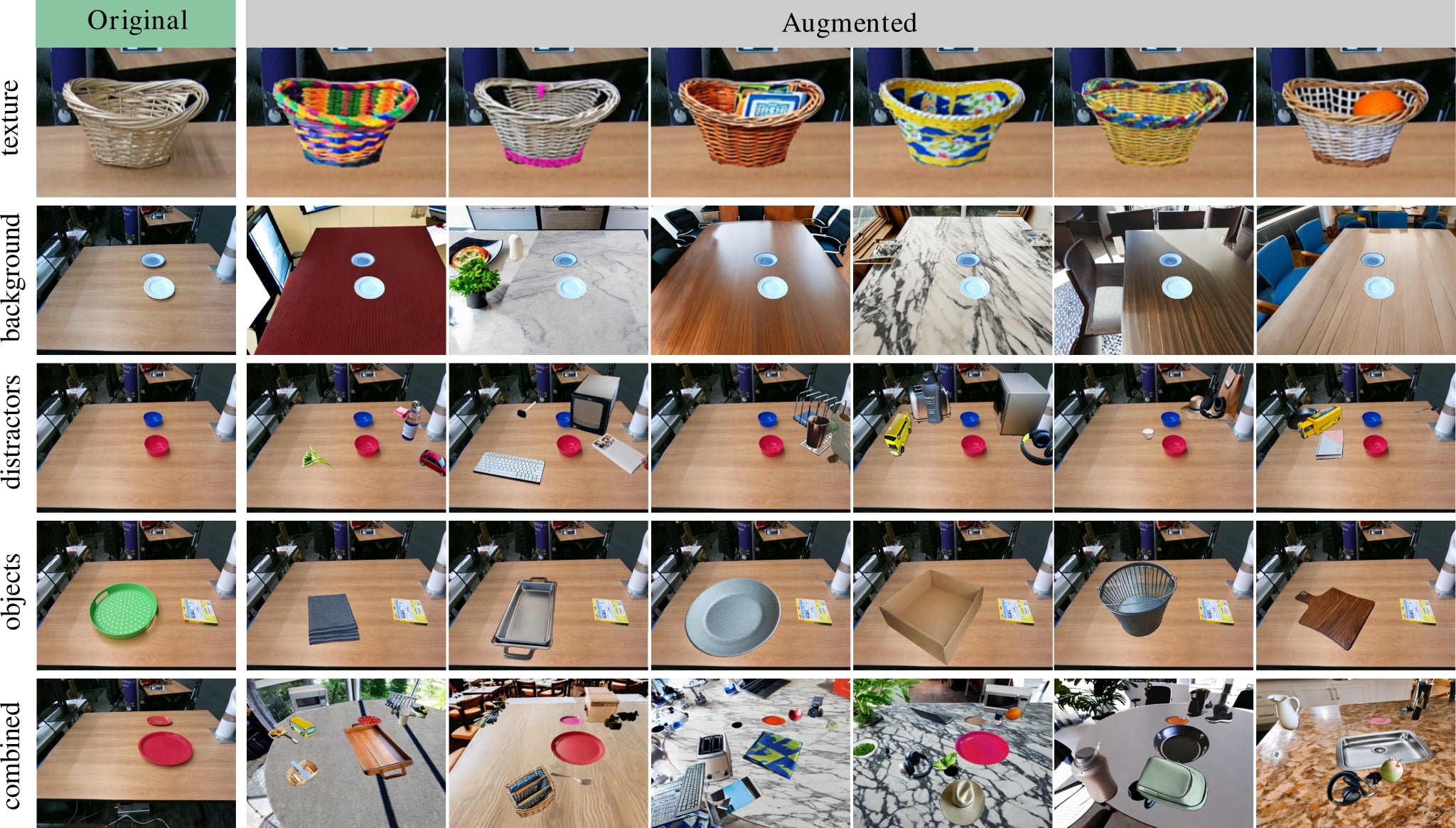}
    \caption{Our framework provides the ability to augment the scene by changing the object texture (first row), changing the background (second row), adding distractors (third row) and changing object categories (fourth row)}
\label{fig:controllabe_augmentation}
\end{figure*}
\begin{figure*}[!h]
    \centering
    \includegraphics[width=\textwidth]{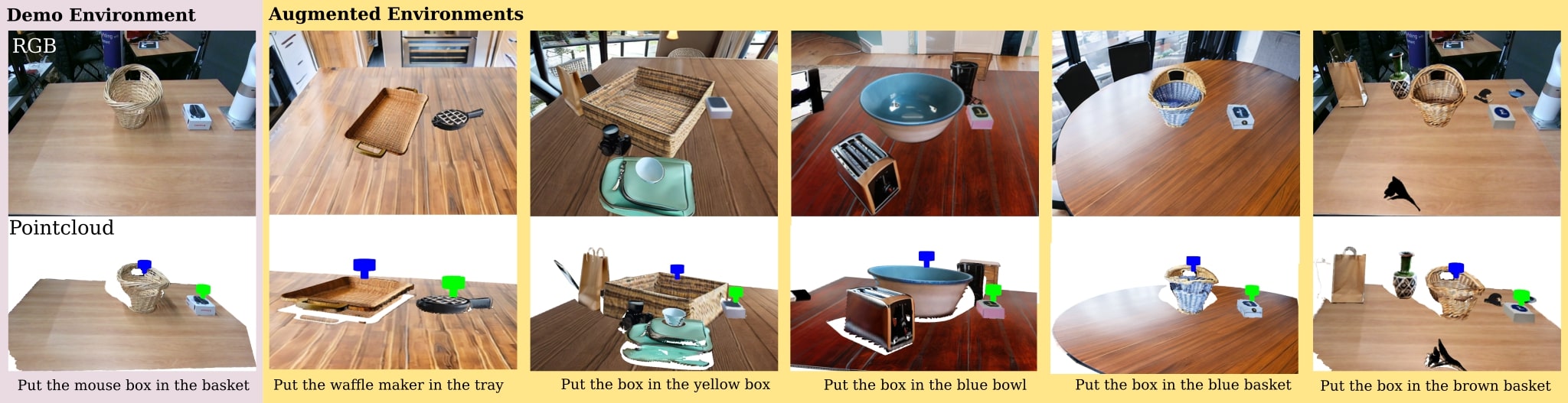}
    \caption{In the low-data regime, our semantic augmentation is controllable and allows more
physically plausible augmentations. We augment the scene in both the RGB and depth information while preserving visual coherence between the RGB and depth modalities. This approach enhances the versatility of the augmentation pipeline, making it suitable for a wide range of methods that utilize RGBD data format.}
    \vspace{-1em}
    \label{fig:augmented_scene_RGBD}
\end{figure*}
In this setting, we focus on how to perform controllable augmentation that is structure-aware and physically plausible on both RGB and depth images. By maintaining consistency between the augmented RGB images and their corresponding depth maps, our framework enables the development of more robust and generalizable algorithms that can effectively leverage the complementary information provided by both modalities.

Given a task on a tabletop, our goal is to perform data augmentations on the visual appearance and 3D geometry of 1) the object being grasped or the target receptacle, 2) distractor objects 3) the table background. Creating a new scene directly on 2D image space often ignores the physical plausibility in an uncontrolled way with no regard for functionality, which is unlikely to retain the semantic invariance that we desire. To appropriately retain semantic invariance, we propose a more controlled image generation scheme. In particular, we assume access to masks $\mathcal{M}(o)$ for every observation $o$, labeling the object of interest and the target receptacle. To generate a diversity of visuals, we consider augmentation both ``in-category" and ``cross-category", as described below:

\textbf{In-category augmentation}
We define in-category augmentation as augmenting objects that are in the same category such as swapping texture. For an in-category generation, we take the provided mask $\mathcal{M}(o)$ of the object to grasp (or the target receptacle) and the original RGB image, and apply a pre-trained depth-ware image conditioned text-to-image diffusion model \citep{rombach2021highresolution} to generate novel visual appearances for objects from the same category. 
Given that the generative model uses the original image as input we use randomly generated novel text prompts to create greater visual diversity. Since visual appearance is strongly correlated with the color and material of an object we ensure our text prompts involve these properties. For instance, we use different colors such as red, orange, and yellow, and materials such as glass, marble, and wood.
Importantly, since the same object masks are used with different prompts, the resulting positions and underlying 3D geometry of the scene remain the same, thus ensuring semantic invariance. 

\textbf{Cross-category Augmentation}
While in-category generation provides a degree of visual diversity, it often falls short of generating novel objects and backgrounds. To encourage more diverse augmentation, we must consider replacing object categories altogether and augmenting background scenes. 
To replace the original object $O_i$ (e.g. a basket) with a new object of a different category (e.g. a bucket) we can naively use model inpainting to generate images directly over the masked object (similar to in-category augmentation).
However, given shape, size, and geometric differences between object categories such simple inpainting will often result in incorrect images since the inpainting model does not guarantee geometric consistency. This is problematic for robotic manipulation where the underlying geometry of the scene is important for robot action.

\begin{figure}[!h]
    \centering
    \includegraphics[width=0.49\textwidth]{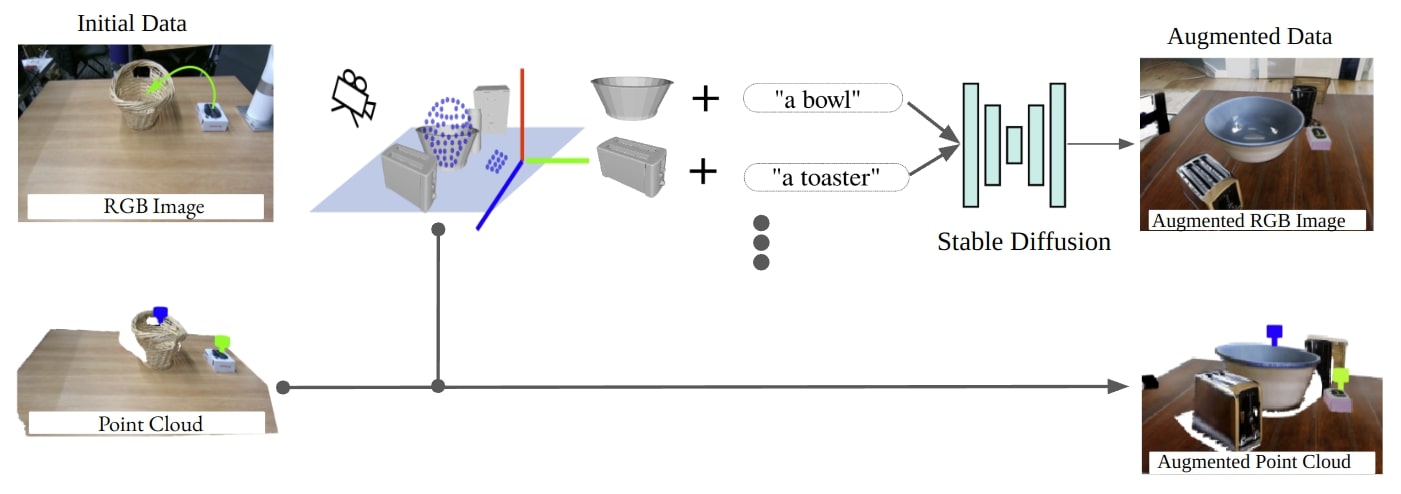}
    \caption{We leverage 3D object assets and simulation, and use text-to-image diffusion models to generate a visually realistic appearance while updating the original depth map, resulting in geometric consistent augmentation}
    \vspace{-0.5em}
    \label{fig:rgbd_augmentation_pipeline}
\end{figure}

To maintain 3D consistency and create physically plausible augmentations we instead use a dataset of object meshes. Specifically, shown in Figure \ref{fig:rgbd_augmentation_pipeline}, we load object meshes from different categories into the scene and render images using the same camera pose from the original data collection. We use the new object mesh with a depth-aware diffusion model\citep{rombach2021highresolution}, as described for an in-category generation, to create new visual scenes. To augment background scenes with distractor objects $D_i$, we randomly choose a new object mesh from a family of object assets and render it on the table. We compute collisions by checking for overlapping bounding boxes (in image space) between the generated distractor $D_i$ and masks $\mathcal{M}(o)$ for the object to grasp and the target receptacle and remove this distractor if it is in a collision. In this way, the simulation ensures 3-D consistency and physical plausibility, while the generative model allows for significant visual diversity. 

\textbf{Language Augmentation} One benefit of leveraging a text-to-image generative model is to automatically generate corresponding language descriptions for the new augmented scenes. For the demonstration of "Put the apple in a box", when replacing the category of the original object $O_i$ labeled with the description $T_i$ (e.g. "a box"), with a new object  $O_j$ with the text prompt $T_j$ (e.g. "a plate"), we can automatically augment the original language description of "Put the apple in a plate" for the generated scene. 

We visualize examples of In-category augmentation and Cross-category augmentation in Figure \ref{fig:controllabe_augmentation}, and compare the consistency between the augmented RGB and depth in Figure \ref{fig:augmented_scene_RGBD} 

So far, we have talked about how to combine simulation 3D object assets and manual object masks for generative models to augment visual scenes while reserving 3D geometry and physical plausibility, and how we can train a language-conditioned robot policy. However, the assumption of having 3D assets and manually labeled masks makes it less scalable for larger robot datasets such as video trajectories. Next, we are going to introduce how to use generative models for scalable augmentation in robot video trajectories which are fully automatic. 

\subsubsection{Scalable Augmentation for Multi-Task Data}
In order to augment large multi-task datasets at scale, we develop an automatic augmentation strategy that doesn't require any manually specified parameters such as object masks or object meshes and doesn't require training or fine-tuning any model. 


Starting with an initial collection of robotic behaviors, we bootstrap this dataset by generating multiple semantically augmented versions of it, while keeping the robot's behavior consistent in each trajectory. These semantic alterations are produced by implementing frame-by-frame augmentations within each trajectory. 
that is fully automatic. In particular, we implement two types of scene augmentations on RGB images:
Object Augmentation: Using the robot's joint angles in a specific trajectory frame, we apply forward kinematics to derive both the robot's mask and the position of its end-effector. The end-effector's location is used to guide SegmentAnything  \citep{sam} in generating a mask for the object being manipulated. This object is then altered through inpainting, based on textual prompts. To ensure temporal consistency, we employ TrackAnything  \citep{trackanything} to track the object across the trajectory.
Background Augmentation: Segment Anything \citep{sam} is utilized to select a group of background objects that don't intersect with either the robot's mask or the mask of the interacting object. We then inpaint these background areas using an overall mask created from the aggregation of all object masks identified by Segment Anything. This approach allows for varied background alterations in the scene.

\begin{figure}[!h]
    \centering
    \includegraphics[width=0.45\textwidth]{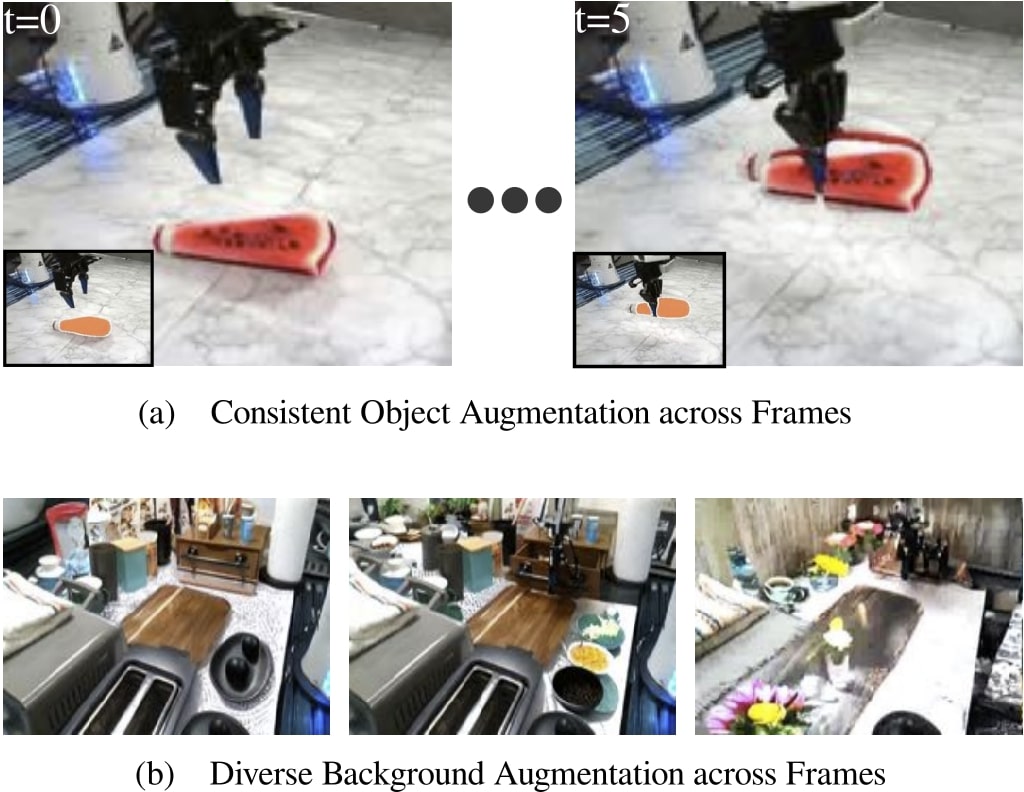}
    \caption{Scalable augmentation in unstructured environment from video trajectories. (a) We use the location of the endeffector to prompt SAM \citep{kirillov2023segment} and get interaction object mask for inpainting, and keep it consistent across video frames using TrackAnything \citep{trackanything}. Images inside the black box show the original frame with the object mask predicted by SAM. (b) We track the masks for the robot and interaction objects, and randomly inpaint regions in the background returned by SAM, resulting in diverse background augmentation across frames.}
    \vspace{-1em}
    \label{fig:auto_aug}
\end{figure}

\subsection{Policy Learning}
In this section, we first explain how our augmentation pipeline can benefit single-task policies that rely on RGBD data, then show how to extend our policy learning to multi-task domains more efficiently with larger-scale datasets.

\begin{figure*}[!h]
\vspace{-0.5em}
    \centering
    \includegraphics[width=0.9\textwidth]{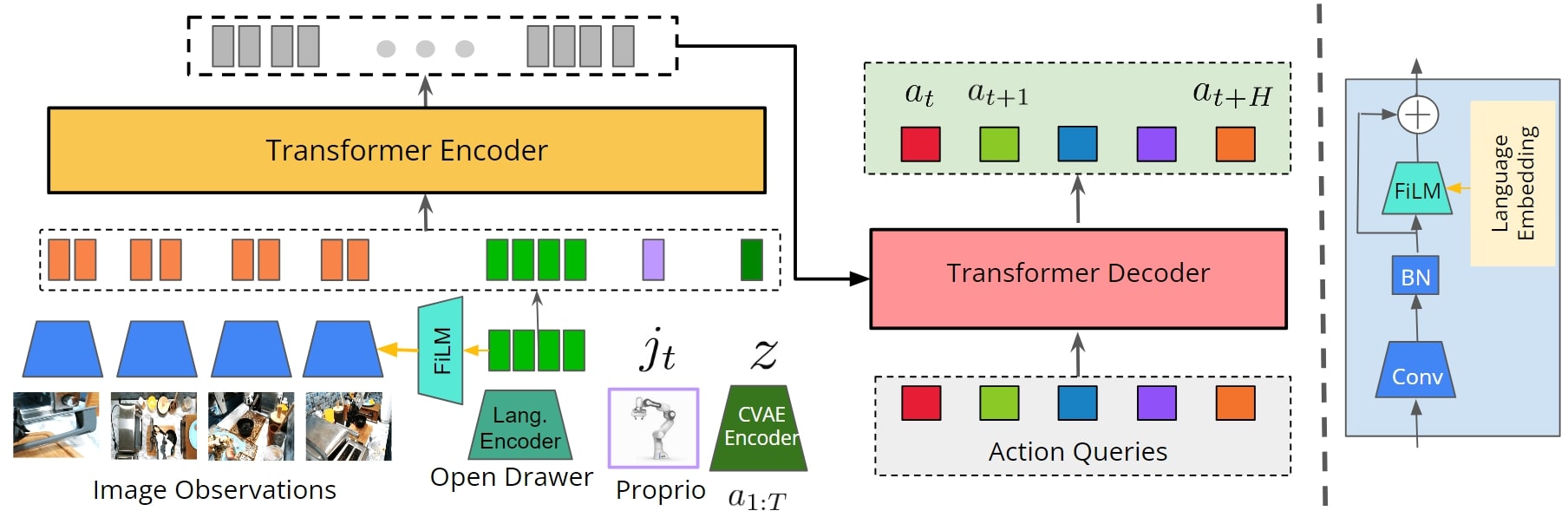}
    \vspace{-0.8em}
    \caption{Policy architecture for Multi-Task Policy that is instantiated as a CVAE whose decoder is a Transformer. The language encoder receives as input a language description of the task, and we provide four different views of the scene at each time step as the observations.}
    \vspace{-1em}
    \label{fig:test_scenes}
\end{figure*}
\subsubsection{Single-Task Policy Learning}
Our single-task learning network is based on Transporter Network \citep{zeng2020transporter} and CLIPort \citep{shridhar2021cliport}. The transporter network learns spatial correspondences between visual features, enabling sample efficiency for visual policy learning. CLIPort aims to combine language understanding with robotic manipulation. CLIPort combines the CLIP model with a Transporter Network. CLIP provides the ability to connect visual features with language descriptions, while the Transporter Network provides efficient spatial generalization capability for table-top pick and place rearrangement tasks. We incorporate the augmented language prompts together with new observations to train a CLIPort which takes language and RGBD observations and predicts pick and place locations. 

\subsubsection{Multi-Task Policy Learning}
To develop a generalizable manipulation policy with a reasonable data budget, we need an efficient policy architecture. Our goal is to train the policy to stay close to nominal behaviors in scenarios that are within the training distribution, and also be generalization to test-time variation and new task contexts while exhibiting smooth temporally correlated behaviors. 

Our policy framework is based on a Transformer model, as described in~\citep{vaswani2017attention}, which has sufficient capacity to process multi-modal, multi-task robotic datasets.  To handle data multi-modality issues, we follow prior works~\citep{act} and integrate a CVAE~\citep{kingma2013vae} that encodes action sequences into latent \textit{style} embeddings $z$. The CVAE decoder which is based on transformers is conditioned on these latent embeddings $z$.  We refer to this CVAE decoder (which takes latent $z$ as input) as our transformer policy.

This approach of treating the policy as a generative model is particularly effective for adapting to the diverse nature of teleoperation data. It ensures that important trajectories that are crucial for precision yet potentially stochastic are not overlooked.
To handle multi-task data of robot trajectories, we integrate a pre-trained language encoder~\citep{gadre2022clip} that outputs an embedding $\mathcal{T}$ of a particular task description.
Instead of solely predicting next-step actions, we also predict actions $H$ steps in the future.
Further, instead of simply discarding the prior predicted actions, we temporally aggregate overlapping actions (from previous predictions) when executing action at each step~\citep{act}.
This approach has two advantages. First, predicting temporally extended actions helps reduce the compounding error. Second, using temporal-aggregation allows for smooth actions which is important for high-frequency control.
Since many of our scenes can have large occlusions (especially when the robot reaches close to an object), we use multiple cameras (4 camera views) in our setup. Our policy uses a transformer ~\citep{vaswani2017attention} of sufficient capacity to handle multi-modal multi-task robot datasets.

At time-step $t$, the transformer encoder takes four camera views, $o^{1:4}_t$, the joint pose of the robot $j_t$, the style embedding from the CVAE $z$, and the language embedding $\mathcal{T}$. 

We utilize FiLM-based conditioning, as detailed by~\citep{perez2018film,rt1}, to ensure that the image tokens effectively concentrate on the language instructions. This focus is crucial to prevent any confusion in the policy about the task at hand, especially in scenarios where multiple tasks are feasible within the same scene. The processed tokens are then passed to the decoder of the Transformer policy, which is equipped with fixed-position embeddings. This setup allows the decoder to generate the upcoming set of actions (comprising $H$ actions) for the current timestep.
Overall, our proposed architecture extends ACT~\citep{act} to multi-task ACT using appropriate language conditioning. Since the demonstration dataset contains diverse skills across tasks, we show that the VAE prior can capture such behavior diversity. Finally, we demonstrate for the first time that action-chunking and temporal aggregation are useful for learning diverse multi-task behaviors for quasi-static (low-frequency control) tasks in diverse scenes.

\section{System Setup}
\begin{figure}[!h]
\vspace{-1em}
    \centering
\includegraphics[width=0.49\textwidth]{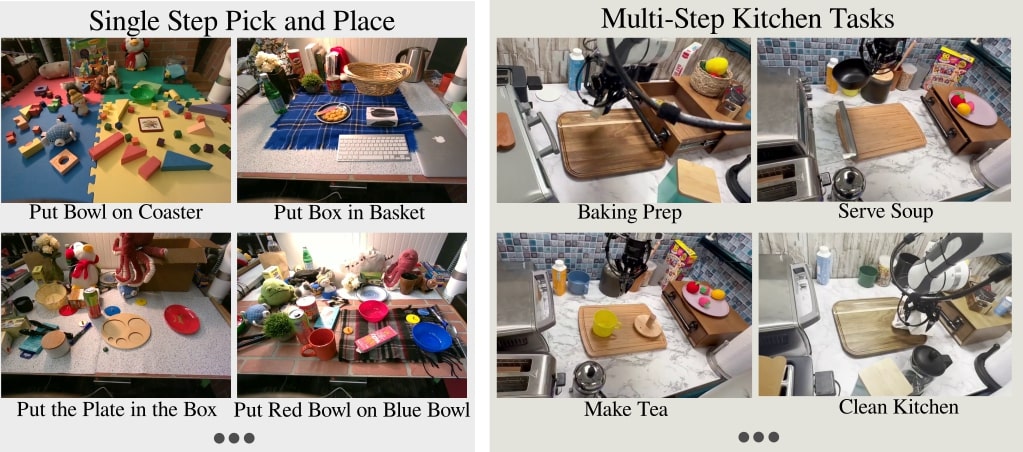}
    \caption{We show the effectiveness of our framework in two groups of tasks. Single-step Pick-and-Place tasks aim to demonstrate generalization across entirely different cluttered environments. Multi-step Kitchen tasks additionally show data efficiency on complex behavior cloning tasks.}
    \vspace{-1.3em}
    \label{fig:task_overview}
\end{figure}
\subsection{Task Overview}
\begin{figure}[!h]
    \centering
    \includegraphics[width=0.49\textwidth]{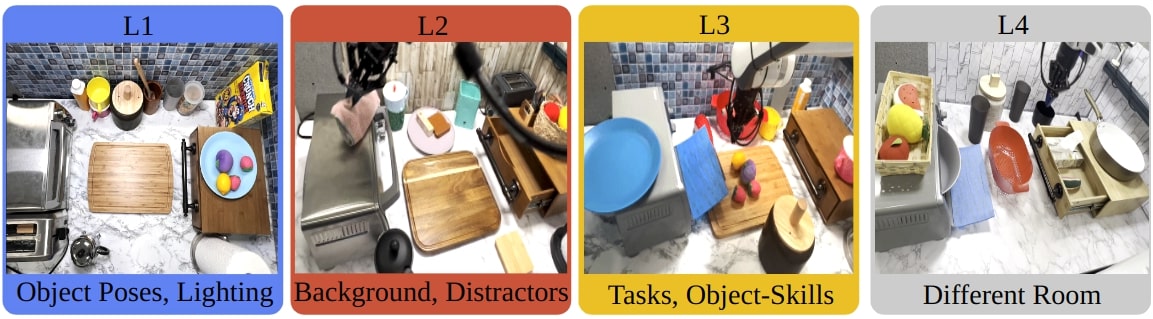}
    \caption{We consider 4 levels of generalization. L1: unseen object poses and lighting variation. L2: unseen background distractors. L3: new tasks and objects, L4: different rooms.}
    \vspace{-0.5em}
    \label{fig:generalization_levels}
\end{figure}
To demonstrate the effectiveness of generative augmentation, We consider evaluations at different generalization levels by applying randomization on a scene. In particular, we define 4 types of unseen environments to evaluate how well generative augmentation can help with generalization at unseen situations. \textbf{L1(Effectiveness)}: The agent's ability to generalize across changes in the placement and orientation of objects, as well as varying lighting conditions. \textbf{L2 (Robustness)}: Adaptability to new backgrounds, diverse variations of distractor objects, and the presence of previously unseen distractor objects in the scene. \textbf{L3 (Generalization)}: Capability to handle entirely new tasks, encompassing novel combinations of objects and skills. \textbf{L4 (Strong Generalization)}: generlization in adapting to different room environments. See Figure \ref{fig:generalization_levels}.

Toward this end, we define two groups of tasks. (1) Single Step Pick-and-Place Tasks that aim to demonstrate strong generalization in cluttered environments and entirely different environments and objects (L3 and L4). (2) Multi-step Kitchen Tasks which aim to demonstrate the effectiveness of generative augmentation in complex tasks and skill learning from video trajectories (L1, L2, L3 and L4). Figure \ref{fig:task_overview} shows the overview of our tasks. 

\subsection{Generative Augmentation}
As described in Semantic Data Augmentation, we present two types of generative augmentation. 

The \textbf{Structure-Aware Augmentation} generates RGBD augmentation which requires object 3D meshes to generate cross-category augmentations and distractors. To perform this augmentation, we use 40 object meshes from the GoogleScan dataset \citep{downs2022google} and Free3D \citep{free3d}. Of these, we choose 11 objects to augment the original object of interest and 12 objects to augment the target receptacle. Any of the remaining 38 objects are then randomly chosen as distractors. During augmentation, we randomly select which components (table, object texture, shape, distractors) to change to generate the augmented training dataset. For each demonstration, we apply augmentation 100 times resulting in 1000 augmented environments per task. The augmented data is then passed into Cliport \citep{shridhar2022cliport} to learn a language-conditioned policy for Pick-and-Place tasks.

\textbf{Scalable Augmentation} as described in Scalable Augmentation for Multi-Task Data, improves the augmentation efficiency of the structure-aware method, which is automatic and does not require any manual effort in specifying masks, object meshes etc. Since this type of augmentation only operates on RGB images, we train a transformer-based visual policy that only takes RGB observations.

\subsection{Real-world Setup}
\begin{figure}[!h]
    \centering
    \includegraphics[width=0.49\textwidth]{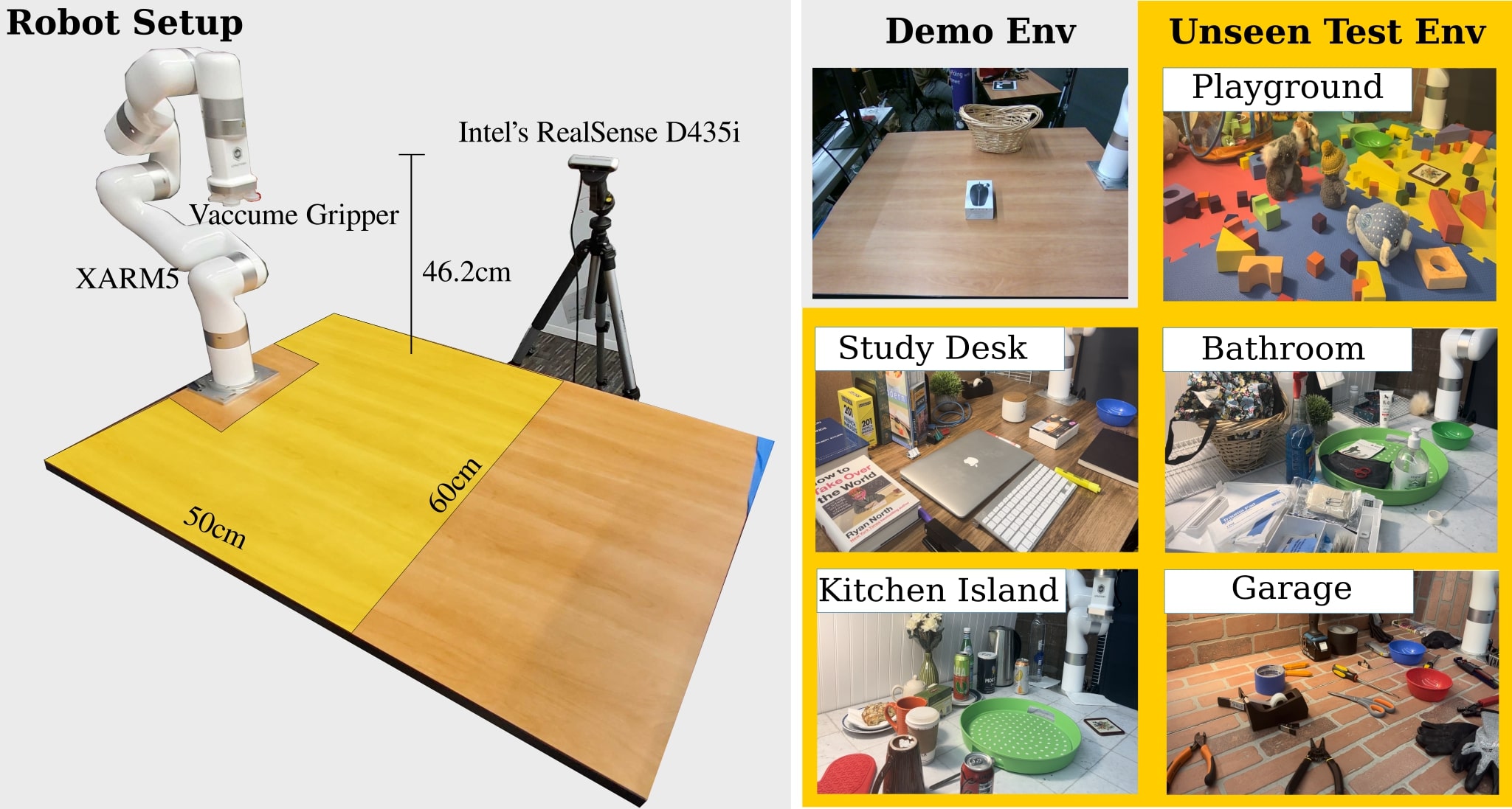}
    \caption{real-world setup and examples of test environments used for pick-and-place tasks .}
    \label{fig:pickplace_setup}
\end{figure}

\begin{figure}
    \centering
\includegraphics[width=0.49\textwidth]{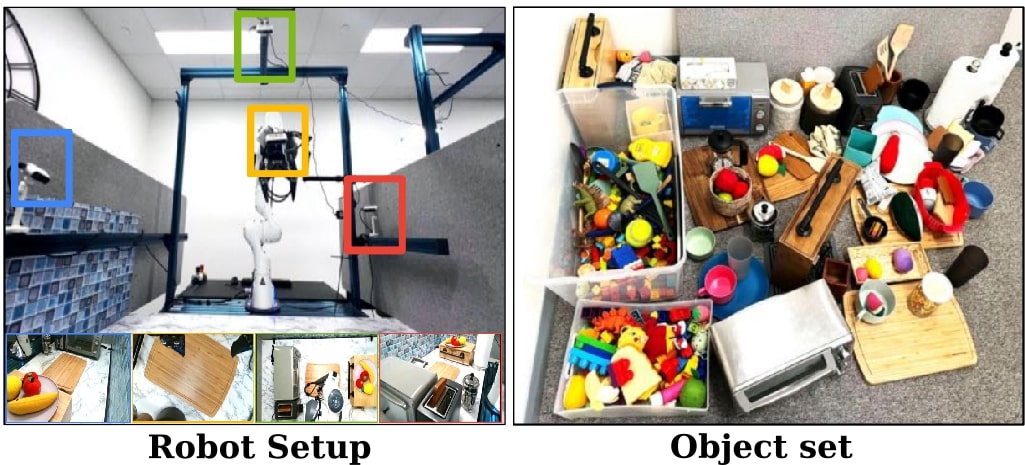}
    \caption{Real-world setup for multi-task kitchen tasks and objects used in the experiments.}
    \label{fig:roboagent_hardware}
\end{figure}
For the Pick-and-Place tasks, we use the 6 DoF xArm5 with a vacuum gripper manipulator and control it directly in the end-effector space. 
Figure~\ref{fig:pickplace_setup} shows our overall setup. We use an XArm mounted on a table in a well-lit room. 
We use a tripod with a depth camera mounted on top and position the tripod such that it has a full view of the robot as well as any objects on the table.
We use an automated setup to collect demonstrations for our pick-place task. Specifically, we use the image captured by the frontal camera and project it into a 2D top-down image and height map. We use this image and have users annotate pick-and-place locations on it. We then convert these pick-and-place image coordinates to world coordinates and use an inverse kinematics controller to reach these positions and perform the pick-place gripper actions.
Overall, we collect data for 10 tasks and for each task we collect 10 demonstrations.
All data is collected in one single environment as shown as ”Demo Environment” in Figure \ref{fig:pickplace_setup}. Appendix A provides further details on each task.

To guide the robot to complete the new tasks in the cluttered environment, we largely build on the architecture and training scheme of CLIPort \citep{shridhar2022cliport}, which combines the benefits of language-conditioned policy learning with transporter networks \citep{zeng2020transporter} for data-efficient imitation learning in tabletop settings. 
CLIPort \citep{shridhar2022cliport} requires RGBD observations as input, which is obtained from an Intel RealSense Camera (D435i). We then manually label the object masks for the collected demonstrations and apply structure-aware augmentation to generate a larger dataset of augmented RGBD data. The input observations are then projected to a top-down view which CLIPort takes, together with language prompts, and predict where to pick and place \citep{zeng2020transporter}. 
\begin{figure*}[!h]
    \centering
    \includegraphics[width=\textwidth]{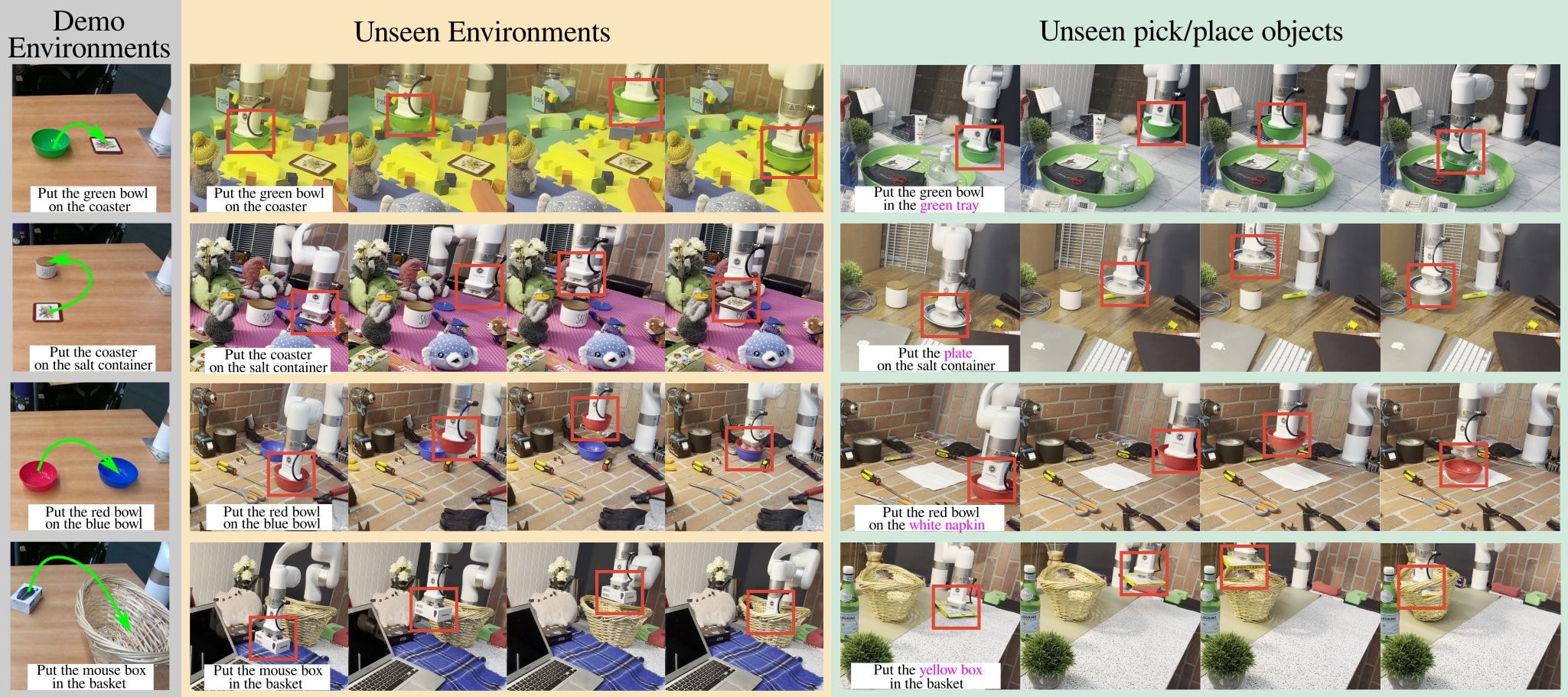}
    \caption{Examples of real-world experiments on Pick-and-Place tasks using single-task CLIPort \citep{shridhar2021cliport}. Given demonstrations in one simple environment, our augmentation framework diversifies the training dataset and enables the robot to generalize unseen environments and objects.}
    \vspace{-1em}
    \label{fig:real_world_genaug_qualitatives}
\end{figure*}
Going beyond simple Pick-and-Place tasks, we additionally conduct experiments on multi-task kitchen environments. Figure \ref{fig:roboagent_hardware} shows our overall setup. We use a kitchen setup that consists of common everyday objects, a Franka Emika Panda arm with a two-finger Robotiq gripper fitted with Festo Adaptive Fingers. We mount three static cameras (top, left, right) to provide a full view of the scene. We further use a wrist camera mounted for more precise motions. Our multi-camera setup provides an exhaustive view of the workspace, which allows for robust policy learning.

We collected 7,500 trajectories using teleoperation by a human operator over two months. All the trajectories are collected in various kitchen-like settings, utilizing a Franka Emika robot \citep{haddadin2022franka}. The teleoperation setup, based on the system described in \citep{kumar2015mujoco}, was operated with VR controllers. This dataset encompasses a wide range of manipulation skills, including activities such as opening and closing drawers, pouring, pushing, dragging, picking up, and placing objects, among others, all involving a variety of everyday items. Examples of tasks are shown in Figure \ref{fig:task_overview}.
\subsection{Baseline Definition}

\begin{figure}[!h]
\vskip -1.2em 
    \centering
    \includegraphics[width=0.49\textwidth]{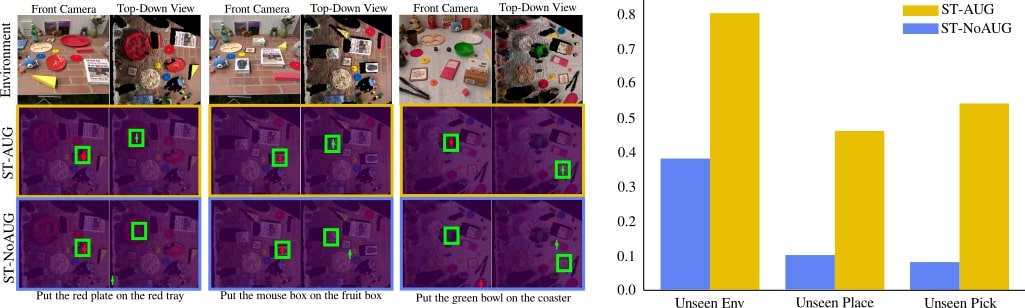}
    \caption{Real-world experiments for single step pick-and-place tasks using \citep{shridhar2021cliport}. We observe almost 40\% improvement on average across 10 different tasks, shown on the right. Qualitative comparisons are visualized on the left, where green boxes represent ground truth, red markers represent prediction on pick locations and green markers represent place predictions. More detailed per-task comparison can be found in Appendix C}
    \vspace{-1em}
    \label{fig:real_world_genaug_baselines}
\end{figure}
Shown in Figure \ref{fig:real_world_genaug_baselines} and Figure \ref{fig:main_result}, we apply our augmentation method to single pick-and-place tasks and name this as \textbf{ST-AUG} and compare it against training without augmentation \textbf{ST-NoAUG}. Similarly, we name experiments with our augmentation for multi-step kitchen tasks as \textbf{MT-AUG} compared to training without \textbf{MT-NoAUG}. 

Additionally, in our multi-task experiments, we evaluate against various baselines that employ visual policy learning in robotics:
\textit{Single Task Agents}: We assess policies based on ACT~\citep{act} that are trained for specific individual tasks and evaluated on those same tasks. Since these policies do not require generalization across different tasks and scenes, they serve as a rough benchmark or "oracle" for task-specific performance.
\textit{Visual Imitation Learning (VIL)}: We compare our approach with standard multi-task visual imitation learning that is conditioned on language.
\textit{CACTI~\citep{cacti}}: this baseline is the previous multi-task learning framework that also incorporates some scene augmentations to aid generalization.
\textit{RT1~\citep{rt1}}: We also implement and test an agent similar to RT1 as another baseline.
\textit{BeT~\citep{bet}}: We adapt the Behavior Transformer (BeT) architecture, add language conditioning, and train it for multi-task purposes.

\section{Experiments}
\begin{table*}[!h]
\centering
\caption{Real-World Robot Experiments tested on 10 tasks. On average, our framework achieves $85\%$ success rate on unseen environment, $52\%$ on the unseen object to place, and $45\%$ on the unseen object to pick.}
\label{table:genaug_robot_experiment}
\begin{tabular}{P{1.9cm}|P{0.9cm}|P{0.9cm}|P{0.9cm}|P{0.9cm}|P{0.9cm}|P{0.9cm}|P{0.9cm}|P{0.9cm}|P{0.9cm}|P{0.9cm}|P{0.9cm}} 
\hline
                   & \begin{tabular}[c]{@{}c@{}}bowl to \\Coaster\end{tabular} & \begin{tabular}[c]{@{}c@{}}box to \\basket\end{tabular} & \begin{tabular}[c]{@{}c@{}}bowl to \\bowl\end{tabular} & \begin{tabular}[c]{@{}c@{}}plate \\to tray\end{tabular} & \begin{tabular}[c]{@{}c@{}}box to \\ tray\end{tabular} & \begin{tabular}[c]{@{}c@{}}plate \\to box\end{tabular} & \begin{tabular}[c]{@{}c@{}}plate to \\ plate\end{tabular} & \begin{tabular}[c]{@{}c@{}}coaster \\to salt\end{tabular} & \begin{tabular}[c]{@{}c@{}}coaster \\to pan\end{tabular} & \begin{tabular}[c]{@{}c@{}}box to \\ box\end{tabular} &
                   \begin{tabular}[c]{@{}c@{}}\textbf{Average}\end{tabular} \\ \hline
Unseen Env & 0.8                                                        & 0.9                                                      & 1                                                                & 1                                                            & 1                                                      & 0.9                                                     & 0.9                                                                  & 1                                                                    & 0.5                                                            & 0.5     & 0.85                                                        \\ \hline
Unseen Place       & 0.7                                                        & 1                                                        & 0.5                                                              & 0.3                                                          & 0.6                                                    & 0.3                                                     & 0.4                                                                  & 0.4                                                                  & 0.4                                                            & 0.6      &0.52                                                         \\ \hline
Unseen Pick        & 0.2                                                        & 0.6                                                      & 0.5                                                              & 0.6                                                          & 0.7                                                    & 0.3                                                     & 0.3                                                                  & 0.7                                                                  & 0                                                              & 0.6     &0.45                                                          \\ \hline

\end{tabular}
\end{table*}

We evaluate the effectiveness of our framework in both the real world and simulation. Our goal is to: (1) demonstrate our framework is practical and effective for real-world robot learning, (2) compare our method with other baselines in end-to-end vision manipulation tasks. We will first show our results in a real-world setting for both single-task learning and multi-task learning, followed by an in-depth analysis and discussions. In addition, all simulation results are detailed in Appendix B.

\subsection{Real-World Experiments}
We conduct a thorough evaluation of the efficacy of our approach focusing on two different learning settings. We first evaluate its performance in simple pick-and-place tasks using single-task language-conditioned policies. This evaluation serves as a benchmark for the fundamental capabilities of our framework in a low-data regime. We then extend our experiments beyond the pick-and-place tasks, aiming to demonstrate the generalization and adaptability of our method. In particular, we show how our approach performs when learning multi-task policies for behavior cloning tasks with video trajectories. 
\begin{figure*}[!h]
    \centering
    \includegraphics[width=\textwidth]{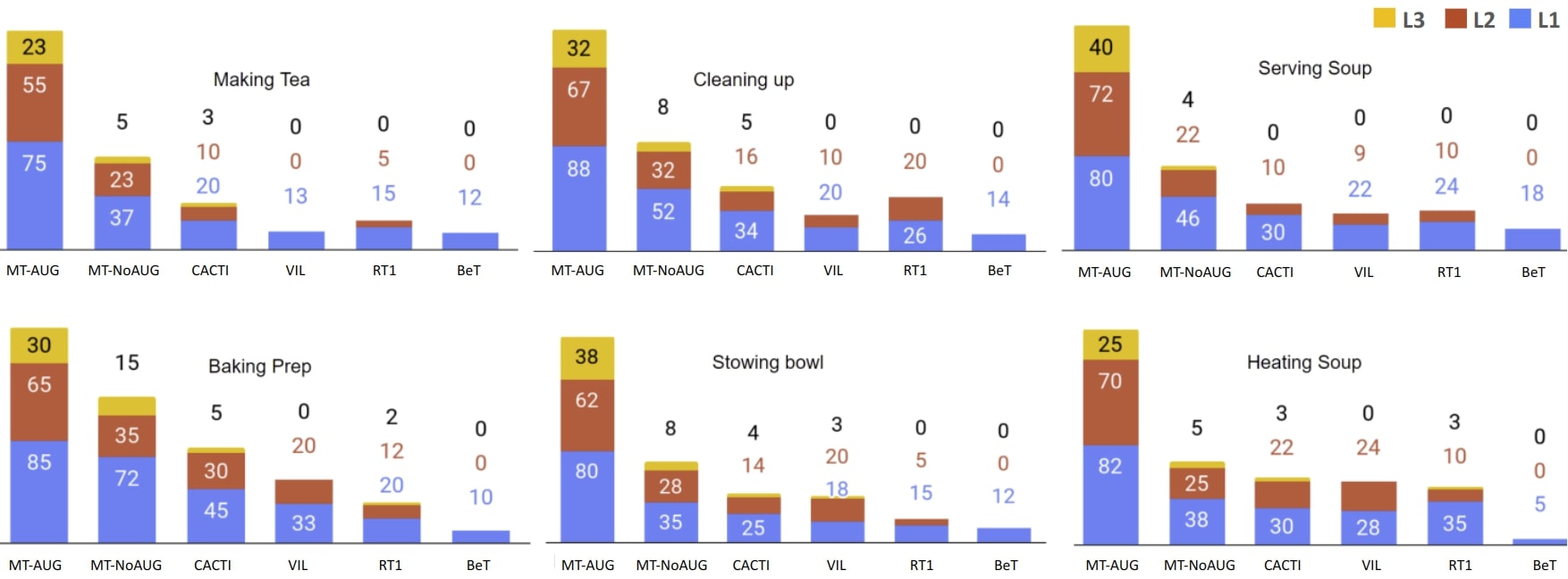}
    \caption{We compare training MT-AUG with and without generative augmentations, as well as compare training MT-AUG networks and other
baselines, for different activities, with L1, L2, L3 levels of generalization.}
    \vspace{-0.5em}
    \label{fig:main_result}
\end{figure*}
\begin{figure*}[!h]
    \centering
    \includegraphics[width=0.99\textwidth]{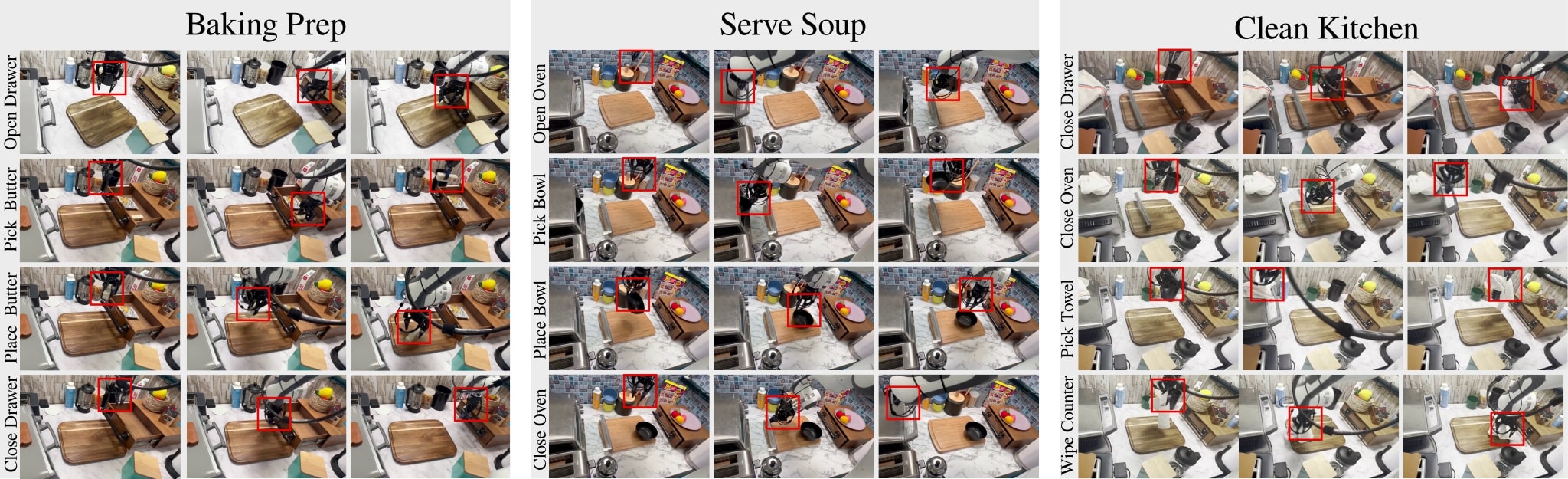}
    \caption{Qualitative results of rollout in complex multi-task kitchen tasks, showing results for tasks under "Baking Prep", "Serve Soup" and "Clean Kitchen" activities.}
    \label{fig:roboagent_qualitatives}
\end{figure*}
\subsubsection{Pick-and-Place (L3 and L4 Generalization)}
To show the generalization capability of a model trained with generative augmentation, we first collect demonstrations of 10 tasks in one single environment and create different styles of test environments such as "Playground", "Study Desk", "Kitchen Island", "Garage" and "Bathroom" as shown in Figure \ref{fig:test_scenes}. For evaluation, we randomly add and rearrange objects from each test style and create unseen environments. Please see Appendix C for further details. 
We train CLIPort with augmented RGBD and text prompts for tasks collected in the real world and evaluate in various unseen environments. In particular, for each task, we randomly choose an environment style from Figure \ref{fig:test_scenes}, randomly rearrange and add objects on the table to create 10 unseen environments, 10 scenes with unseen objects to pick, and 10 scenes with unseen objects to place. We observe that our approach shows a significant generalization to unseen environments with an average of $85\%$ success rate. On more challenging tasks of unseen objects to pick or place, our method achieves $45\%$ and $52\%$ success rates, which are expected to improve with more demonstrations and more object meshes for augmentation, shown in Table \ref{table:genaug_robot_experiment}.
We compare our approach with CLIPort trained without augmentation, shown in Figure \ref{fig:real_world_genaug_baselines}. To ensure both methods are tested with the same input observations, we evaluate the success rate by comparing the predicted pick and place affordances with ground truth locations. For each task, we evaluate both methods on 5 unseen environments, 5 unseen objects to pick, and 5 unseen objects to place.
We observe generative augmentation provides a notable improvement for zero-shot generalization. In particular, our approach achieves $80\%$ success rate on unseen environments compared to $38\%$ without augmentation. On unseen objects to place, ours achieves $54\%$ success rate compared to $8\%$ without. Finally, ours achieves $46\%$ success rate on unseen objects to pick compared to $10\%$ without. We visualize and compare the differences in their predicted affordances in Appendix C.

\subsubsection{Multi-Task Kitchen Tasks}
\begin{figure}[!h]
\centering
\vspace{-1em}
\includegraphics[width=0.35\textwidth]{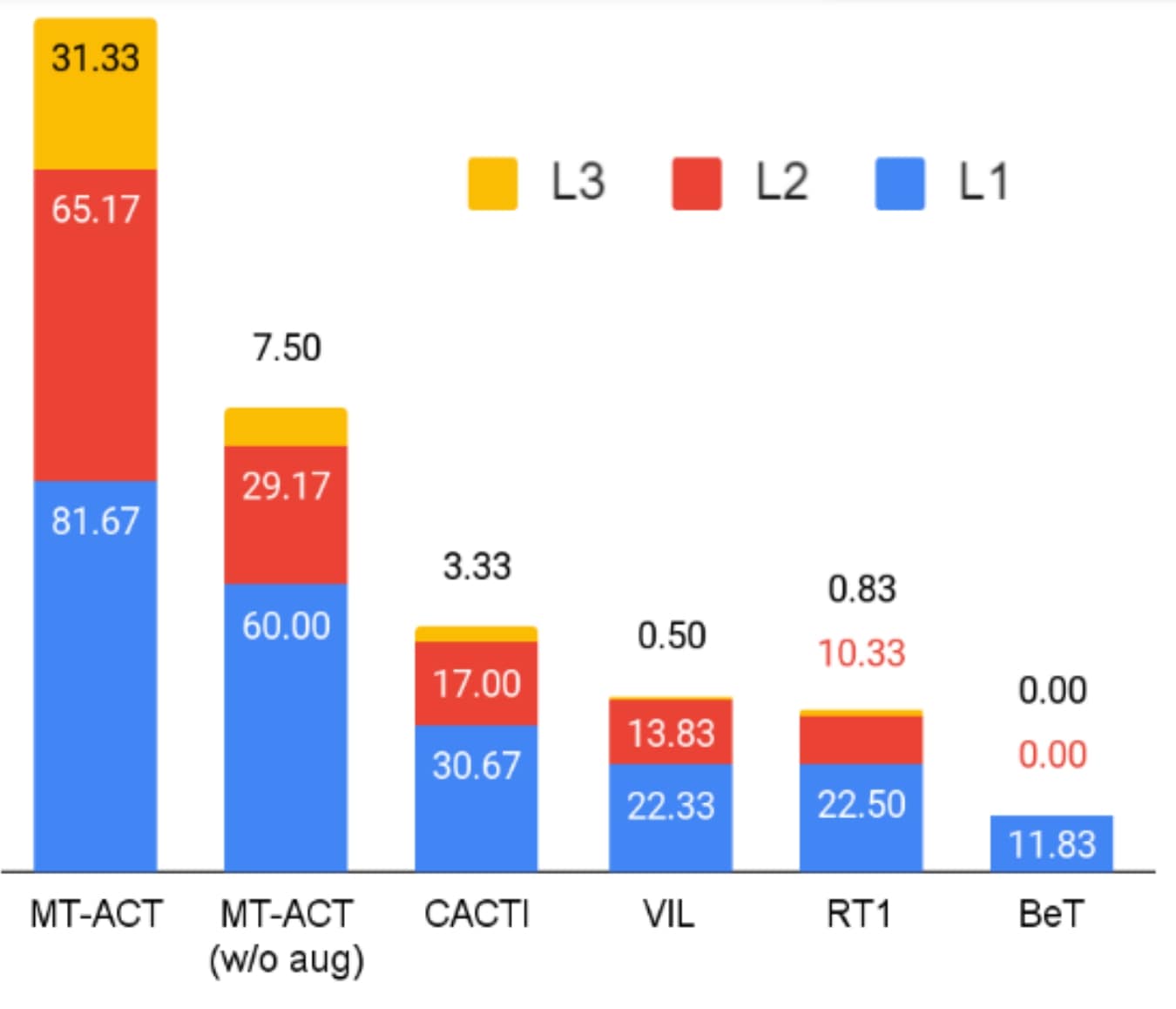}
\vspace{-1em}
\caption{Comparison of Different Multi-Task
(universal policy) results for different levels of generalization}
\vspace{-1em}
\label{fig:mtact_average}
\end{figure}

In this section, we discuss the results of multi-task policy learning experiments, which incorporate automatic video augmentations. The results depicted in Figure~\ref{fig:main_result} span various generalization levels (\textbf{L1}, \textbf{L2}, and \textbf{L3}, see definition in Task Overview) for each activity. Each activity consists of 4-5 tasks, and the results average
over the tasks in an activity. It's important to remember that these generalization levels encompass a range of elements, such as varied table backgrounds and distractors (\textbf{L2}), as well as new combinations of skills and objects (\textbf{L3}). Our findings indicate that our method, due to semantic augmentations and advanced action representations, substantially outperforms all baseline models.
Particularly, as the average results shown in Figure \ref{fig:mtact_average}, while semantic augmentations show a moderate enhancement in L1-generalization (about 30\% relatively), they yield far more significant improvements in L2-generalization (around 100\% relatively) and L3-generalization (approximately 400\% relatively). Given that these augmentations impact both the scenes (including backgrounds and distractor objects) and the target objects being manipulated, they play a crucial role in supporting the policy's ability to adapt to increasingly complex generalization levels.
Furthermore, for some of the more challenging activities like Making-Tea, Stowing-Bowl, and Heating Soup, the boost in performance due to semantic augmentations is notably greater.
Overall, our findings show that traditional visual imitation learning approaches, such as VIL and RT1, which do not utilize augmentations and are trained on a relatively limited dataset, completely fail at L3 and L2 levels. This failure indicates their inability to generalize to novel scenarios, a limitation likely due to the scarcity of data. Additionally, we conducted tests on our policy in an entirely new kitchen environment, replete with novel objects, arrangements, and distractors, essentially testing for L4 generalization.
In this new kitchen setting, across three tasks, we observed an average success rate of 25\% for MT-AUG (with all other baselines achieving 0\%). This demonstrates that even MT-AUG, without semantic augmentations, fails entirely in novel environments, thereby highlighting the significant advantage of our generative augmentation approach for zero-shot adaptation.
\subsection{Ablations}
\begin{figure}[H]
\centering
\vspace{-1em}
\includegraphics[width=0.35\textwidth]{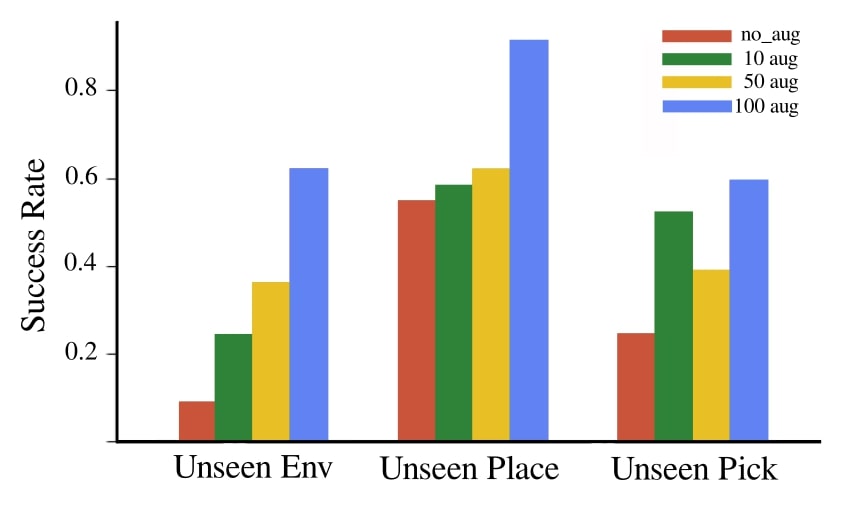}
\vspace{-1em}
\caption{Analysis of the number of augmentation on unseen scenes during pick-and-place tasks}
\vspace{-1em}
\label{fig:genaug_ablation_aug}
\end{figure}
In this section, our goal is to investigate (1) how the number of augmentations affects the generalization performance to unseen environments, and (2) how robust our system is at dealing with disturbance. (3) how reusable our policy is by finetuning on a new task.
\subsubsection{Impact of the number of augmentations} 
We evaluate how the quantity of augmentations affects performance in both pick-and-place tasks and settings involving multi-task tasks. Specifically, in a simulated task like "put the brown plate in the brown box," we experiment with applying augmentations 0, 10, 50, and 100 times. We then evaluate their success rates across three different sets of 100 scenes: those with "unseen environments," "unseen objects to pick," and "unseen objects to place." 

\begin{figure}[H]
\centering
\vspace{-1.2em}
\includegraphics[width=0.3\textwidth]{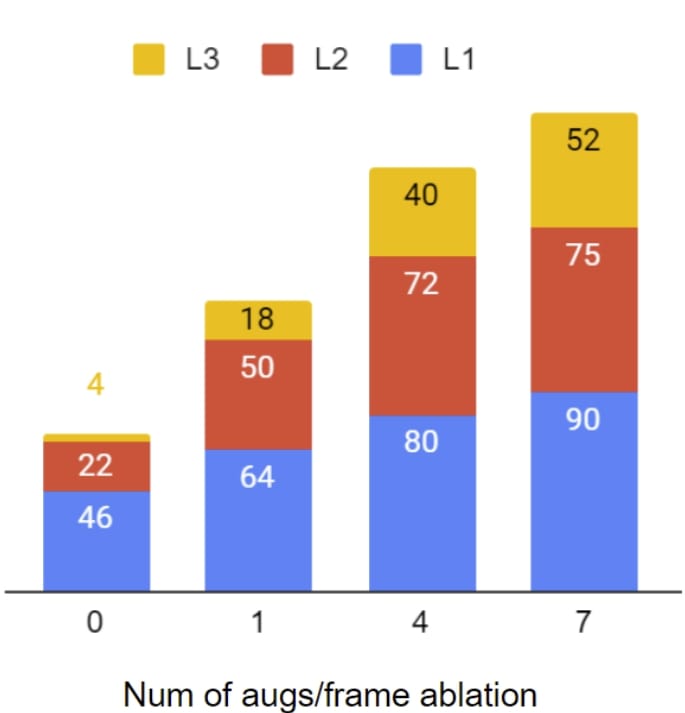}
\caption{Ablation on the number of augmentation per frame in videos for multi-task kitchen tasks}
\vspace{-1.5em}
\label{fig:roboagent_ablation_aug}
\end{figure}

As illustrated in Figure \ref{fig:genaug_ablation_aug} (a), there is a noticeable improvement in performance with an increasing number of augmentations. This indicates the importance of augmentations in enhancing the robust generalization capabilities of the system.
In a multi-task real-world context, we explore the impact of varying the number of augmentations per frame to determine whether a higher number of augmentations contributes to the development of a more effective policy. As shown in Fig.~\ref{fig:roboagent_ablation_aug} (Middle-Right), there is a clear correlation between the frequency of augmentations per frame and the overall improvement in performance. 
These improvements are particularly notable at the L2 and L3 levels, where the policy is expected to generalize to out-of-domain scenarios. 
This boost in performance can be attributed to the introduction of real-world semantic knowledge through the process of data augmentation.

\subsubsection{Robustness Analysis}
We conducted various robustness tests on the universal MT-AUG agent, including manual alterations to the scene during evaluations and introducing system failures such as obstructing views from one, two, or three cameras. We observe that the policy maintains a high level of resilience against these significant active variations. In approximately 70\% of the 20 evaluations conducted for this analysis, the policy successfully accomplished the given task. While the robustness of manual scene alternation might come from the semantic augmentation, the multi-view transformer-based structure in the MT-ACT network can be another factor for the resilience of camera views.

\subsubsection{Plasticity}

\begin{figure}[!h]
\vspace{-1em}
\centering
\includegraphics[width=0.35\textwidth]{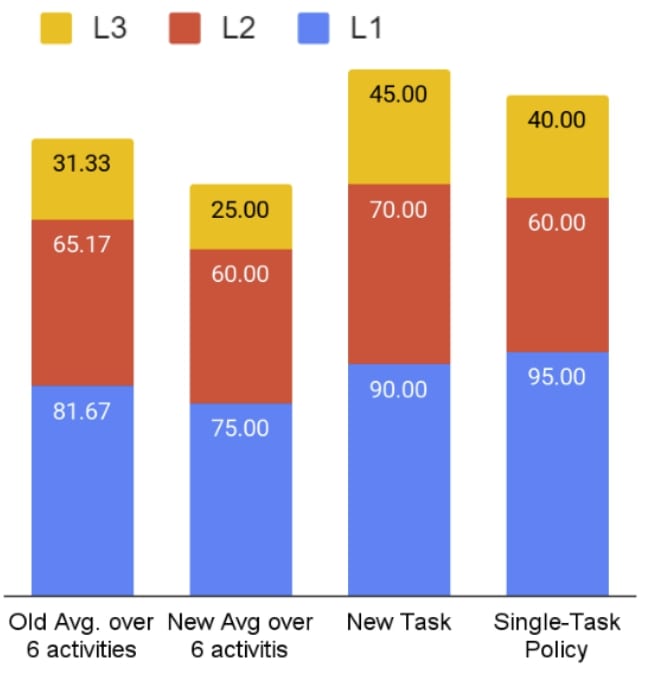}
\vspace{-1em}
\caption{Analysis of the feasibility of fine-tuning MT-AUG for improved deployment by fine-tuning the trained multi-task agent on 50 demonstrations from a new task.}
\vspace{-2em}
\label{fig:reusability}
\end{figure}

In addition, we evaluate the potential of improving the universal MT-AUG agent with new capabilities without necessitating extensive retraining. Starting with the agent already trained on 38 tasks, we proceeded to fine-tune it using a fraction (1/10) of the original data, supplemented with data for an additional, previously untrained task (placing toast in the toaster oven). This new task comprised 50 trajectories, each expanded with 4 augmentations per frame, resulting in a total of 250 trajectories. As observed in \ref{fig:reusability}, the fine-tuned agent successfully learns the new task without any notable decline in its performance on the original 6 activities. Moreover, it shows marginally better performance in L2 and L3 generalization compared to a single-task policy trained solely on augmented data for the new task.  This suggests the efficient reusability of data in our approach. 

\section{Discussion}
\subsection{Trade-off between structure consistency and scalability}
\begin{figure}[!h]
    \centering
    \includegraphics[width=0.49\textwidth]{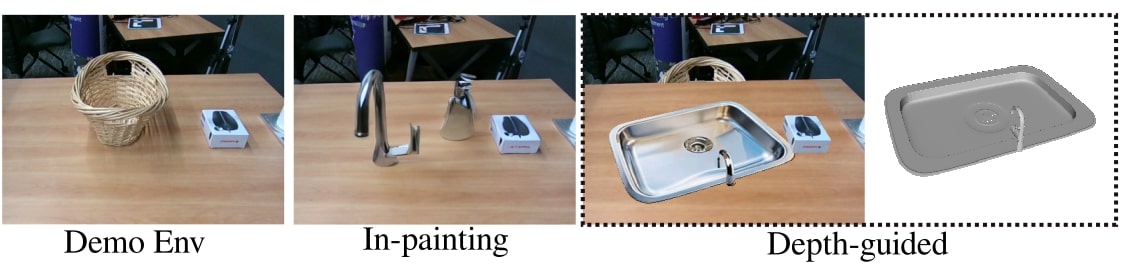}
    \vspace{-1.5em}
    \caption{Trade-off between structure consistency (Depth-guided) and scalability (in-painting).}
    \vspace{-1em}
    \label{fig:trade_off}
\end{figure}
We present two ways of doing generative augmentation: (1) structure-aware augmentation uses depth-aware diffusion models for low-data regimes that require access to 3D assets while (2) scalable augmentation uses in-painting diffusion models which enable automatic augmentation in videos and bigger datasets. We observe the inpainting models might result in less structural augmentation while depth-guided augmentation conditioned on 3D geometry yields more physically plausible augmentation. We visualize the difference between these augmentations in figure \ref{fig:trade_off}.

\subsection{Failure Cases}
\subsubsection{Failures in Generative Augmentation}
\begin{figure}[!h]
    \centering
    \includegraphics[width=0.49\textwidth]{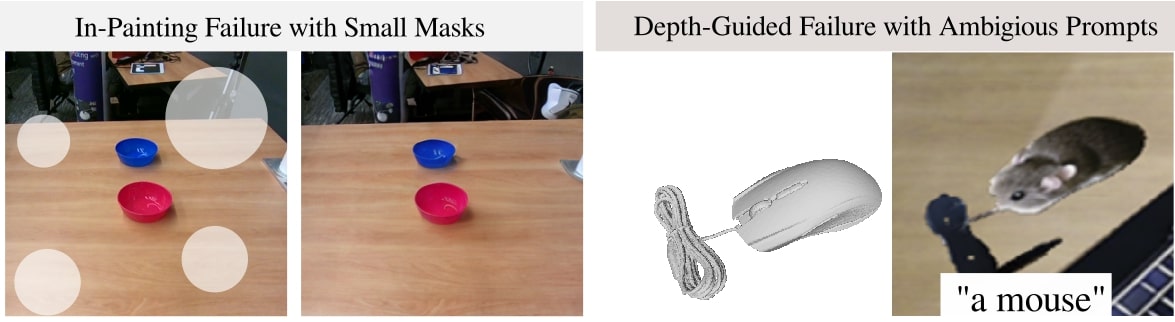}
    \vspace{-1.5em}
    \caption{Failure Cases in generative augmentation with in-painting diffusion models and depth-guided diffusion models.}
    \vspace{-1.5em}
    \label{fig:augmented_failure}
\end{figure}

We observe two typical failure modes during generative augmentation. Shown in Figure \ref{fig:augmented_failure}, (1) When applying in-painting diffusion models, smaller mask regions usually lead to invalid augmentation. (2) depth-guided augmentation results to unrealistic augmentation when the prompt is not specific, such as "a mouse" instead of "a computer mouse". 

\subsubsection{Failures in Robot Experiments}
We observe failure cases usually occur when the background color is similar to the pick or place object. Or one of a few distractors has a very bright color or similar color. We expect this can be improved by increasing the number of augmentations in the training set, such that the training data can have higher coverage of possible combinations of the scenes. For the multi-task experiments, we observe the failure of MT-AUG when the skill required at test time is different from any of the skills in the original teleoperation dataset. This is because our generative augmentations only target visual changes in the scene and cannot augment actions with completely different behaviors. 
\begin{figure}[!h]
    \centering
    \includegraphics[width=0.45\textwidth]{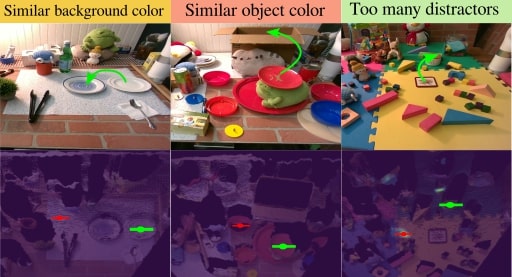}
     \vspace{-0.5em}
    \caption{Failure Cases in Real-World Robot Experiments.}
    \vspace{-2.3em}
    \label{fig:augmented_scene}
\end{figure}
\section{Limitations}
\textbf{  Action Assumption} Despite showing promising visual diversity, our work does not augment action labels and reason about physics parameters such as material, friction, or deformation, thus it assumes the same action still works on the augmented scenes. For the augmented cluttered scenes, we assume the same action trajectory is not colliding with the augmented objects. 
\textbf{Augmentation Speed} It usually takes about 30 seconds to complete all the augmentations for one scene, which might not be practical for some robot learning approaches such as on-policy RL. 
\textbf{Long-horizon Learning}
An important constraint in our work is the focus solely on isolated skills for each task. A promising avenue for future research lies in devising methods that can autonomously compose these skills. This development would be crucial for tackling tasks that require extended planning and execution horizons.
\section{Conclusion and Future Work}
We present Generative Augmentation, a novel system for augmenting real-world robot data. Our approach leverages a data augmentation approach that bootstraps a small number of human demonstrations into a large dataset with diverse and novel objects. By automatically growing an initial small robotics dataset using semantic scene augmentations, we train a language-conditioned policy using the augmented dataset and demonstrate that generative augmentation can enable a robot to generalize to entirely unseen environments and objects. For future work, we are interested in developing a more scalable augmentation approach that is consistent and fast while still maintaining physical plausibility. Additionally, whether a combination of language models and vision-language models can yield impressive scene generations would be a promising direction. Our framework only augments robot data on its visual appearance, another interesting extension of our work is to do augmentation on the action level, by leveraging recent video generation approaches like Unisim \cite{yang2023learning} and inferring inverse dynamics from generated video frames.

\bibliographystyle{SageH.bst}
\bibliography{scripts/references}




\clearpage
\newpage
\appendix
\section*{Appendix A: Augmented Dataset in Simulation}
\begin{figure}[!h]
    \centering
    \includegraphics[width=0.49\textwidth]{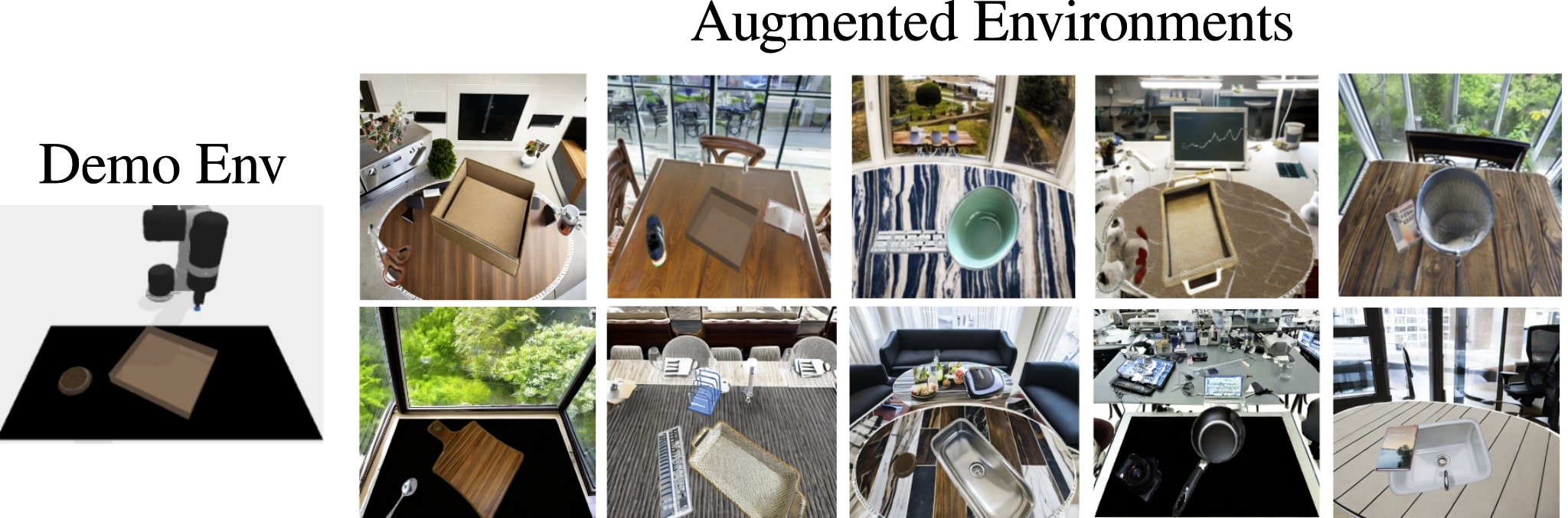}
    \vspace{-1em}\caption{Augmented dataset for demonstrations collected in simulation.}
    \vspace{-1em}
    \label{fig:sim_aug}
\end{figure}

Given demonstrations from a task collected in simulation, we apply augmentation 100 times for each demonstration. We visualize examples of the augmented dataset in Figure \ref{fig:sim_aug}. 

\begin{figure}[!h]
    \centering
    \includegraphics[width=0.49\textwidth]{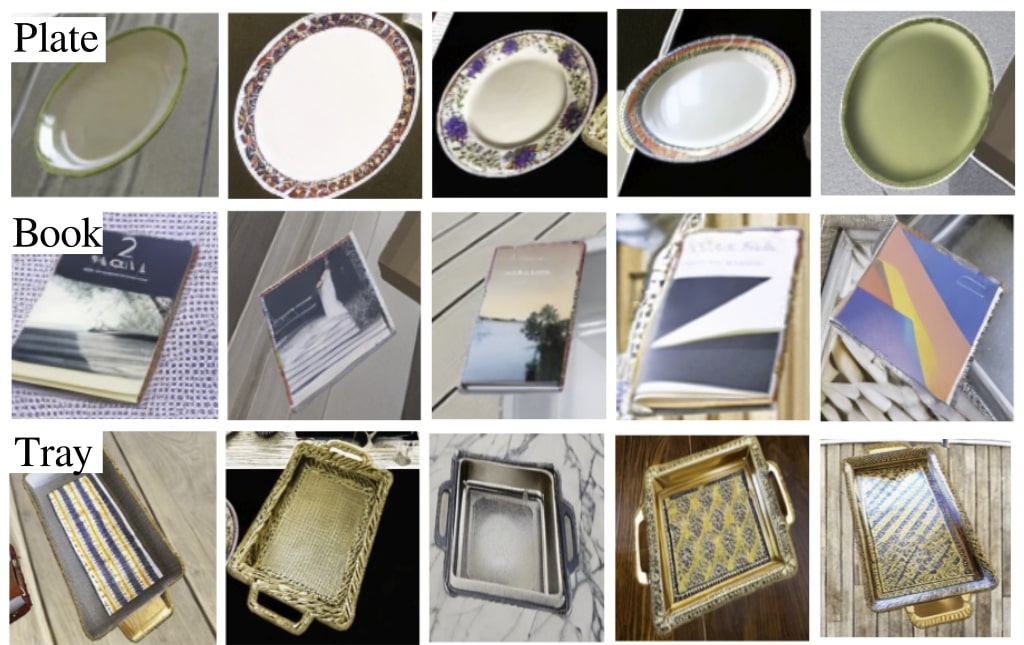}
    \vspace{-1.5em}\caption{Diversity of the appearance of the generated objects}
    \vspace{-1.5em}
    \label{fig:diverse_aug}
\end{figure}

We also observe diverse visual augmentation on the same object template, as shown in Figure \ref{fig:diverse_aug}. Given different text prompts, our method is able to generate different and realistic textures.

\section*{Appendix B: Simulation Experiments}
\subsubsection{Single-Task Pick and Place Tasks}

\subsection{Simulation}
\subsubsection{Pick-and-place environment}
To further study the effectiveness of our work, we conduct in-depth experiments with other baselines focusing on pick and place tasks in simulation. In particular, we organize baseline methods as (1) in-domain augmentation methods and (2) learning from out-of-domain priors, as described below.

\textbf{In-domain augmentation methods}
(1) "No Augmentation" does not use any data augmentation techniques. 
(2) "Spatial Augmentation" randomly transforms the cropped object image features to learn rotation and translation equivariance, as introduced in TransporterNet \citep{zeng2020transporter}. 
(3)"Random Copy Paste" randomly queries objects and their segmented images from LVIS dataset \citep{gupta2019lvis}, and places them in the original scene. This includes adding distractors around the pick or place objects or replacing them. Further visualization of this approach can be found in Appendix C. 
(4)"Random Background" does not modify the pick or place objects but replaces the table and background with images randomly selected from LVIS dataset.
(5)"Random Distractors" randomly selects segmented images from LVIS dataset as distractors.

\textbf{ Learning from out-of-domain priors}
In addition, we investigate whether learning from a pretrained out-of-domain visual representation can improve the zero-shot capability on challenging unseen environments. In particular, we initialize the network with pre-trained R3M \citep{nair2022r3m} weights and finetune it on our dataset. 

We use baselines described above with two imitation learning methods: TransportNet \citep{zeng2020transporter} and CLIPort \citep{shridhar2021cliport}. Since all the baselines cannot update the depth of the augmented images, we only use RGB images instead of RGBD used in the original TransporterNet and CLIPort. For each baseline, we train 5 tasks in simulation and report their average success rate in Table \ref{table:sim_experiment}. We observe our work notably outperforms other approaches in most of the tasks. One interesting observation is that randomly copying and pasting segmented images or replacing the background images can provide reasonable robustness in unseen environments but are not able to achieve similar performance as ours at unseen objects. This indicates generating semantically meaningful and physically plausible scenes is important. 
\begin{figure*}
    \centering
    \includegraphics[width=0.95\textwidth]{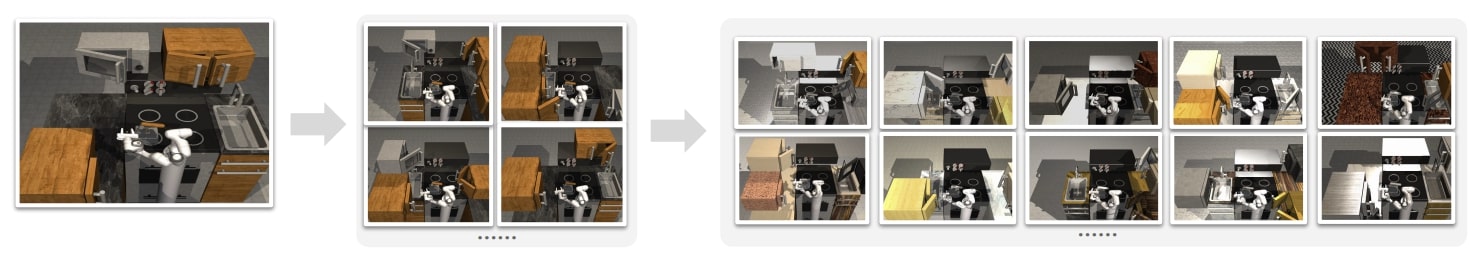}
    \vspace{-1.5em}\caption{For the multi-task kitchen environment in simulation, we randomly change layouts by placing different objects such as the microwave and cabinets in different locations.}
    \vspace{-0.5em}
    \label{fig:random_paste}
\end{figure*}

\begin{table*}
\centering
\small 
\caption{\textbf{Simulation results}. We evaluate the success rates across various training layout randomizations (10, 50, 100) and conduct assessments on both familiar layouts experienced during training and new, heldout layouts not encountered previously. A notable observation is the enhanced ability to generalize to these heldout layouts, which becomes more pronounced with the increase in layout variations introduced during training. This trend highlights the advantages of incorporating robust semantic data augmentations in the Augmentation phase.
}
\begin{tabular}{lcccccccc}
\toprule
   & \multicolumn{2}{c}{Sim $10$}  && \multicolumn{2}{c}{Sim $50$} && \multicolumn{2}{c}{Sim $100$} \\
\cmidrule{2-3} \cmidrule{5-6} \cmidrule{8-9}
   & Train & Heldout && Train & Heldout && Train & Heldout \\ 
\midrule 
State-based & 81.2 +/- 2.2 & 14.1 +/- 3.1 && 83.3 +/- 2.2  & 31.6 +/- 4.4    &&  91.3 +/- 1.1 & 47.2 +/- 4.5   \\ 
Out-domain  & 51.3 +/- 4.3 & 9.5 +/- 2.0  && 64.7 +/- 3.6  & 30.4 +/- 5.6    &&  62.0 +/- 3.8 & 33.1 +/- 4.3   \\
In-domain   & 88.7 +/- 1.2 & 18.8 +/- 2.9 && 72.1 +/- 2.8  & 30.2 +/- 5.2    &&  75.9 +/- 2.6 & 38.4 +/- 4.7  \\
\bottomrule
\end{tabular}
\vspace{-1em}
\label{cacti-sim}
\end{table*}
\begin{figure}[!h]
    \centering
    \includegraphics[width=0.49\textwidth]{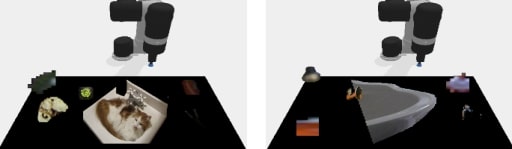}
    \vspace{-1em}\caption{Examples of random copy and paste baseline. We extracted queried segmented images from LVIS dataset and paste them directly on the original demonstration image. This usually leads to low-quality and incomplete image generation.}
    \vspace{-0.5em}
    \label{fig:random_paste}
\end{figure}

\textbf{Visualization of Baseline Data augmentation}
We visualize some examples of randomly copying and pasting segmented images from LVIS dataset \citep{gupta2019lvis}, as shown in Figure \ref{fig:random_paste}.

We observe this baseline often results in unrealistic, low-quality image generation, which does not usually match observations during test time in both the real-world and simulation. 

\subsubsection{Multi-Task Kitchen Tasks}
In addition to simple tabletop pick and place tasks, we also perform in-depth experiments on multi-task kitchen tasks. In this experiment, instead of operating augmentation using generative models, we hope to conduct a simple test and see how much augmentation would improve a multi-task robot policy. We utilize the advantage of simulation and directly perform augmentation to change layout, texture, and lighting conditions.  

\textbf{Task Configuration} The simulation setting for multi-task kitchen tasks comprises 18 distinct tasks involving 8 primary objects, such as activating a light switch, opening a cabinet's left door, and adjusting a knob. Alongside these semantic tasks, we've crafted 100 unique kitchen layouts through randomization. The 8 objects include four stove knobs, a light switch, a kettle, two types of cabinets, and a microwave. We use a standard on-policy RL algorithm, namely NPG~\citep{npg}, to train a fleet of single-task, single-layout expert policies $\pi(s_t)$ from state-based input observations $s_t$. For each task $\mathcal{T}$, we define a reward function $r_{\mathcal{T}}$, and the expert policy $\pi_{\mathcal{T}}$ receives the current simulator state $s_t=\{robot, object\}_{pos, vel}$ as observation at time-step $t$. We generate 100 layout variations for each of the 18 tasks and trains an expert policy for each layout, hence resulting in a total of 900 policies. 
The process of training experts is cost-effective and can be efficiently parallelized. In practical terms, we start a substantial number of training runs simultaneously. Then, we identify and select converged policies as experts by applying a success rate threshold of 90\%.

\textbf{Training Details} We employ a 43-dimensional context vector to inform a versatile, multi-task policy model. A task is considered successful if the manipulated object's final position closely aligns with its intended goal position, within a specified margin of error. This embedding is meticulously designed to encode both the specific pose of the target object for each task and the unique layout configuration. To test the model's adaptability, we introduce 10 new configurations for each task within the simulation environment and vary visual attributes randomly. In particular, for visual representations, we compare with a pre-trained in-domain backbone that is trained with robot simulation data alone, and an out-of-domain backbone trained on internet human videos~\citep{r3m}. For training, we have different levels of simulation layout variations, ranging from 10 to 100 (Sim-10, Sim-50, Sim-100). 

\textbf{Evaluation Details}
During the evaluation, a task is considered successful if the object's final position remains within an error threshold from the target goal position for over five-time steps. This ensures the stability of the object's position, preventing any misclassification of success. To assess the policy's adaptability to new conditions, we introduce an additional 10 layouts for each task in the simulation and vary visual elements such as color, lighting, and texture during the ten evaluation trials. The policy trained with 100 layouts undergoes testing across 5 trials for each of the 1800 task-layout pairings, while the policies with 10 and 50 layouts face 10 trials for each task-layout combination used in training.

Table \ref{cacti-sim} illustrates improved generalization to new layout variations, progressing from Sim-10 to Sim-100. This suggests that incorporating data augmentations with a greater range of layout changes during the training notably enhances the model's ability to generalize beyond its initial domain.

\section*{Appendix C: More Real-World 
Experiments}

\begin{table*}
\centering
\caption{Baseline experiments evaluated in simulation. We compare the average performance of our method with other methods on 5 pick-and-place tasks and observe our work provides a notable improvement at unseen environments and objects.}
\vspace{0em}
\label{table:sim_experiment}
\small 
\setlength{\tabcolsep}{4.5pt} 
\renewcommand{\arraystretch}{1} 
\begin{tabular}{ccccccccccccccccccc} 
\multicolumn{1}{l}{}      & \multicolumn{6}{c}{Unseen Env}                                                        & \multicolumn{6}{c}{Unseen place}                                                           & \multicolumn{6}{c}{Unseen pick}                                                            \\ \cline{2-19} 
\multicolumn{1}{l}{}      & \multicolumn{3}{c}{TransporterNet}            & \multicolumn{3}{c}{CLIPort}                   & \multicolumn{3}{c}{TransporterNet}            & \multicolumn{3}{c}{CLIPort}                   & \multicolumn{3}{c}{TransporterNet}            & \multicolumn{3}{c}{CLIPort}                   \\ \cline{2-19} 
Method                    & 1             & 10            & 100           & 1             & 10            & 100           & 1             & 10            & 100           & 1             & 10            & 100           & 1             & 10            & 100           & 1             & 10            & 100           \\ \cline{2-19} 
No Aug           & 4.8           & 8.1           & 9.8           & 11.7          & 14.3          & 14.4          & 15.1          & 30.4          & 52.6          & 39.4          & 40.8          & 44.6          & 8.5           & 34.6          & 54.9          & 46.0          & 67.0          & 64.1          \\
Spatial Aug      & 11.0          & 12.2          & 8.3           & 23.3          & 16.1          & 26.7          & 44.3          & 50.5          & 65.3          & 26.1          & 36.9          & 50.7          & 53.6          & 57.2          & 66.4          & 38.2          & 56.9          & 80.3          \\
CopyPaste         & \textbf{53.1} & 67.0          & 73.5          & 38.2          & 39.8          & 64.3          & 55.1          & 65.4          & \textbf{84.9} & 39.7          & 55.9          & 73.9          & 48.3          & 67.0          & 76.1          & 52.5          & 65.0          & 81.0          \\
Background         & 53.0          & 75.3          & 79.1          & 33.6          & 62.2          & 55.4          & 24.5          & 22.1          & 35.5          & 7.6           & 9.9           & 17.9          & 44.4          & 40.7          & 35.9          & 19.2          & 52.7          & 72.3          \\
Distractor        & 10.1          & 9.7           & 13.7          & 15.4          & 36.2          & 35.8          & 28.2          & 60.7          & 66.0          & 27.5          & 51.8          & 54.3          & 42.5          & 47.4          & 62.3          & 31.0          & 64.0          & 69.1          \\
R3M              & 4.1           & 6.0           & 4.8           & 22.2          & 16.8          & 20.9          & 43.5          & 40.6          & 41.9          & 30.9          & 43.5          & 57.5          & 45.6          & 45.7          & 41.1          & 46.7          & 50.7          & 72.7          \\
\textbf{Ours}           & 43.9          & 58.5          & 77.6          & 46.6          & 57.0          & 71.9          & \textbf{69.1} & \textbf{76.5} &  83.6    & 62.6          & \textbf{83.9} & \textbf{86.3} & \textbf{75.3} & 75.6          & \textbf{87.2} & \textbf{61.5} & \textbf{77.7} & \textbf{83.1} \\
\textbf{Ours (D)} & 47.8          & \textbf{83.8} & \textbf{91.2} & \textbf{47.2} & \textbf{60.9} & \textbf{73.4} & 39.9          & 67.2          & 74.2          & \textbf{64.8} & 73.8          & 84.6          & 71.2          & \textbf{83.4} & 87.1          & 56.2          & 67.3          & 81.5         
\end{tabular}
\vspace{-0em}
\end{table*}

\begin{table*}[]
\centering
\caption{Evaluating with and without GenAug on unseen scenes collected in the real world across 10 tasks. On average, GenAug shows notable improvement in unseen environments and objects.}
\label{table:robot_baselines}
\begin{tabular}{|l|ccl|ccl|ccl|ccl|ccl|}

\hline
 &
  \multicolumn{3}{c|}{box to tray} &
  \multicolumn{3}{c|}{box to basket} &
  \multicolumn{3}{c|}{coaster to dust pan} &
  \multicolumn{3}{c|}{plate to tray} &
  \multicolumn{3}{c|}{bowl to coaster} \\ \hline
 &
  \multicolumn{1}{c|}{env} &
  \multicolumn{1}{c|}{pick} &
  place &
  \multicolumn{1}{c|}{env} &
  \multicolumn{1}{c|}{pick} &
  place &
  \multicolumn{1}{c|}{env} &
  \multicolumn{1}{c|}{pick} &
  place &
  \multicolumn{1}{c|}{env} &
  \multicolumn{1}{c|}{pick} &
  place &
  \multicolumn{1}{c|}{env} &
  \multicolumn{1}{c|}{pick} &
  place \\ \hline
No GenAug &
  \multicolumn{1}{c|}{0.8} &
  \multicolumn{1}{c|}{0} &
  0 &
  \multicolumn{1}{c|}{0.2} &
  \multicolumn{1}{c|}{0.2} &
  0 &
  \multicolumn{1}{c|}{0.8} &
  \multicolumn{1}{c|}{0.4} &
  0.4 &
  \multicolumn{1}{c|}{0} &
  \multicolumn{1}{c|}{0} &
  0 &
  \multicolumn{1}{c|}{0} &
  \multicolumn{1}{c|}{0} &
  0 \\ \hline
GenAug &
  \multicolumn{1}{c|}{\textbf{1}} &
  \multicolumn{1}{c|}{\textbf{0.6}} &
  \textbf{1} &
  \multicolumn{1}{c|}{\textbf{0.6}} &
  \multicolumn{1}{c|}{\textbf{0.6}} &
  \textbf{0.8} &
  \multicolumn{1}{c|}{\textbf{1}} &
  \multicolumn{1}{c|}{0.4} &
  0.4 &
  \multicolumn{1}{c|}{\textbf{1}} &
  \multicolumn{1}{c|}{\textbf{0.4}} &
  \textbf{0.2} &
  \multicolumn{1}{c|}{\textbf{0.6}} &
  \multicolumn{1}{c|}{\textbf{0.6}} &
  \textbf{0.6} \\ \hline
 &
  \multicolumn{3}{c|}{plate to plate} &
  \multicolumn{3}{c|}{box to box} &
  \multicolumn{3}{c|}{plate to box} &
  \multicolumn{3}{c|}{coaster to salt} &
  \multicolumn{3}{c|}{bowl to bowl} \\ \hline
 &
  \multicolumn{1}{c|}{env} &
  \multicolumn{1}{c|}{pick} &
  place &
  \multicolumn{1}{c|}{env} &
  \multicolumn{1}{c|}{pick} &
  place &
  \multicolumn{1}{c|}{env} &
  \multicolumn{1}{c|}{pick} &
  place &
  \multicolumn{1}{c|}{env} &
  \multicolumn{1}{c|}{pick} &
  place &
  \multicolumn{1}{c|}{env} &
  \multicolumn{1}{c|}{pick} &
  place \\ \hline
No GenAug &
  \multicolumn{1}{c|}{0} &
  \multicolumn{1}{c|}{0} &
  0.2 &
  \multicolumn{1}{c|}{0.2} &
  \multicolumn{1}{c|}{0} &
  0 &
  \multicolumn{1}{c|}{0.6} &
  \multicolumn{1}{c|}{0.2} &
  0 &
  \multicolumn{1}{c|}{0.2} &
  \multicolumn{1}{c|}{0} &
  0.2 &
  \multicolumn{1}{c|}{1} &
  \multicolumn{1}{c|}{0.2} &
  0 \\ \hline
GenAug &
  \multicolumn{1}{c|}{\textbf{1}} &
  \multicolumn{1}{c|}{0} &
  \textbf{0.6} &
  \multicolumn{1}{c|}{\textbf{0.8}} &
  \multicolumn{1}{c|}{\textbf{0.4}} &
  \textbf{0.4} &
  \multicolumn{1}{c|}{\textbf{1}} &
  \multicolumn{1}{c|}{\textbf{0.8}} &
  0 &
  \multicolumn{1}{c|}{\textbf{1}} &
  \multicolumn{1}{c|}{\textbf{0.4}} &
  \textbf{0.4} &
  \multicolumn{1}{c|}{1} &
  \multicolumn{1}{c|}{\textbf{0.4}} &
  \textbf{1} \\ \hline
\end{tabular}
\vspace{-0em}
\end{table*}

\begin{figure}[!h]
    \centering
    \includegraphics[width=0.49\textwidth]{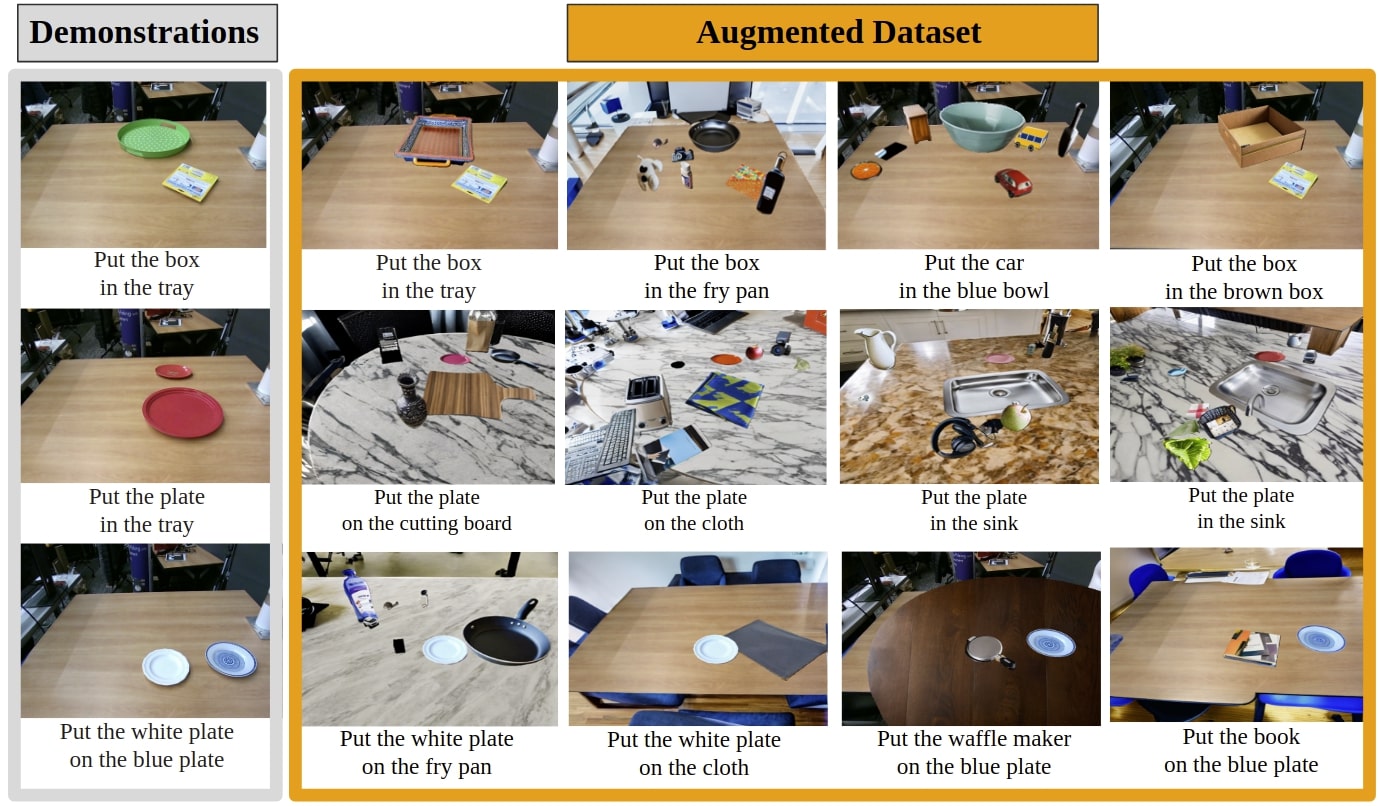}
    \vspace{-1em}\caption{Examples of augmented dataset given observations of demonstrations collected in a simple environment.}
    \vspace{-0em}
    \label{fig:augmented_dataset}
\end{figure}

To further show the effectiveness of GenAug, we compare our approach with CLIPort trained without GenAug, shown in Table \ref{table:robot_baselines}. To ensure both methods are tested with the same input observations, we evaluate the success rate by comparing the predicted pick and place affordances with ground truth locations. For each task, we evaluate both methods on 5 unseen environments, 5 unseen objects to pick, and 5 unseen objects to place.
By averaging the success rates from Table \ref{table:robot_baselines}, we observe GenAug provides a notable improvement for zero-shot generalization. In particular, GenAug achieves $80\%$ success rate on unseen environments compared to $38\%$ without GenAug. On unseen objects to place, GenAug achieves $54\%$ success rate compared to $8\%$ without. Finally, GenAug achieves $46\%$ success rate on unseen objects to pick compared to $10\%$ without. We visualize and compare the differences in their predicted affordances in Figure 13.

\begin{figure*}[!h]
    \centering
    \includegraphics[width=\textwidth]{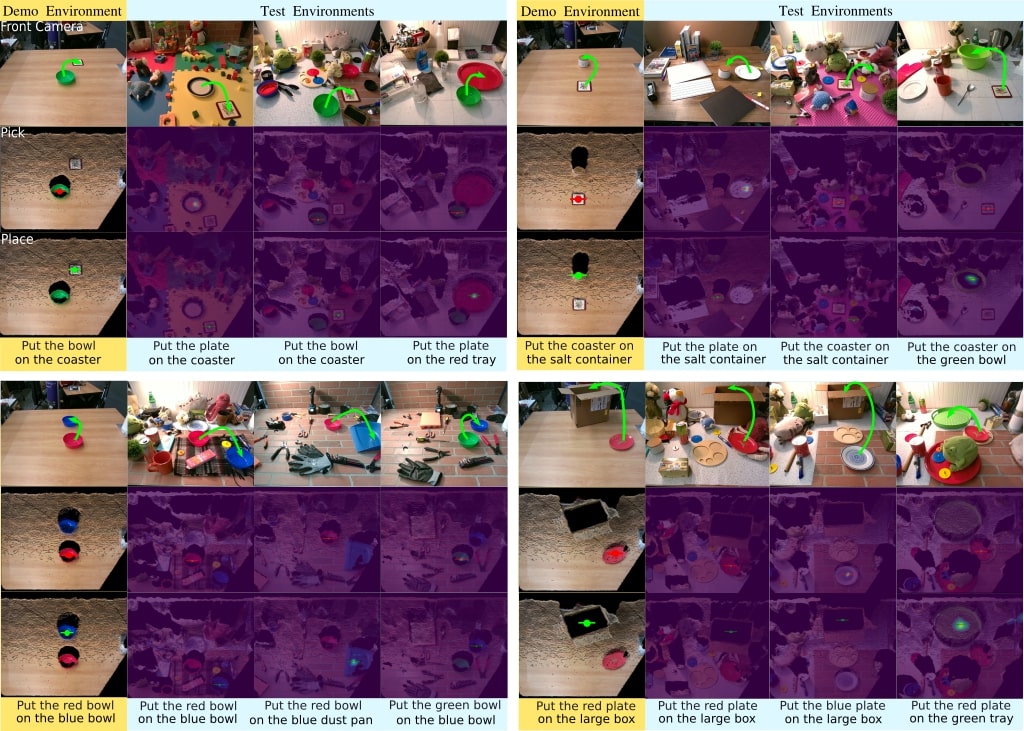}
    \vspace{-0.5em}\caption{Prediction of pick and place locations on various pick and place tasks}
    \vspace{-0.5em}
    \label{fig:robot_experiment}
\end{figure*}

\begin{figure*}[!h]
    \centering
    \includegraphics[width=0.95\textwidth]{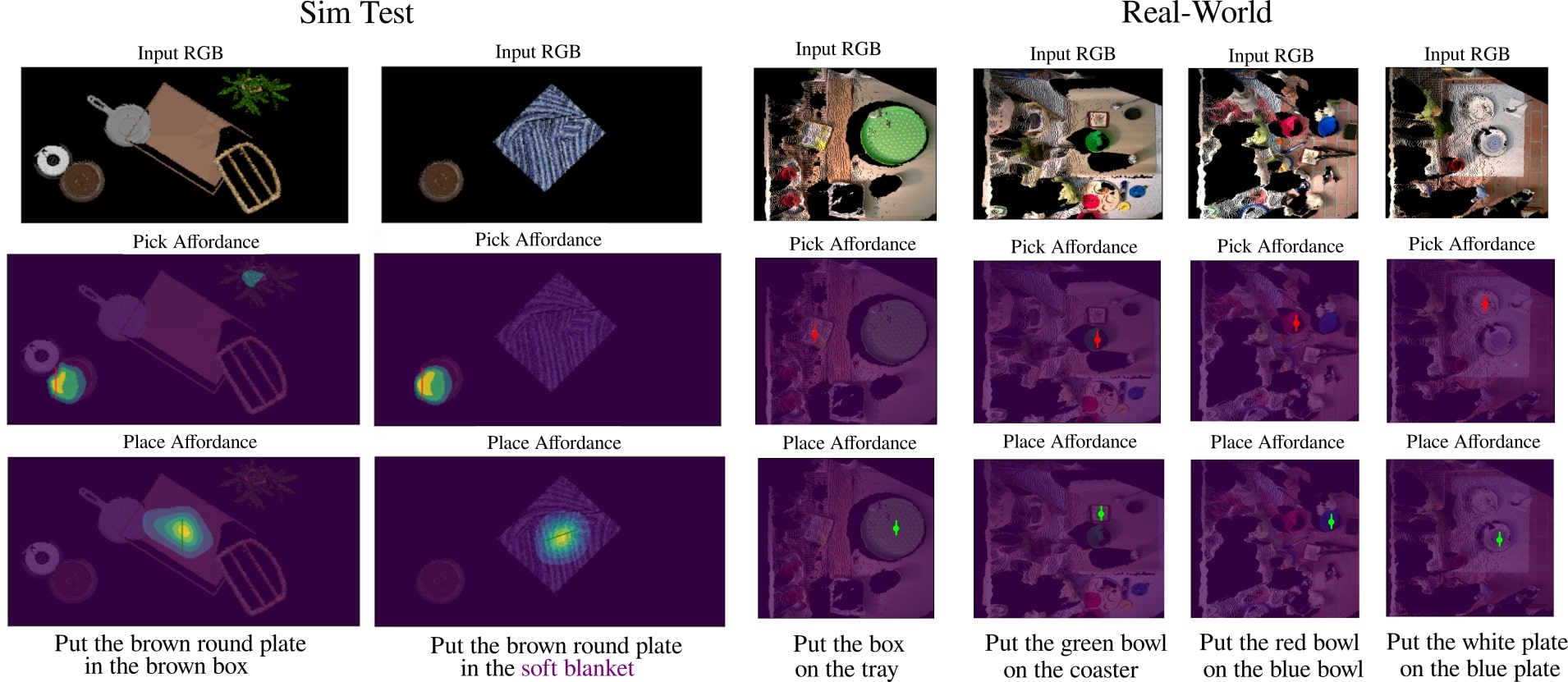}
    \vspace{-0.5em}\caption{Pick and Place affordance predicted by CLIPort that trained on our augmentation framework on unseen environments and objects in simulation and the real world.}
    \vspace{-0.5em}
    \label{fig:affordance}
\end{figure*}

We visualize more results of our pick-and-place real-world experiments. Figure \ref{fig:augmented_dataset} shows more examples of our augmented dataset in the real-world setting. Figure \ref{fig:affordance} shows the affordance map predicted by CLiport at unseen test environments in both simulation and real-world. Figure \ref{fig:robot_experiment} provides more examples of affordance prediction in our pick-and-place real-world experiments.

\section*{Appendix D: Computational Cost}
 For generating the augmented dataset, the average speed for augmenting one mask is about 4 seconds on a 2080Ti GPU, so per-image augmentation is about 10 seconds to 30 seconds depending on the number of masks or distractor objects to augment. Training the overall MT-ACT agent on the augmented dataset takes about 48 hours on a single 2080Ti GPU. Training CLIPort on the augmented dataset takes about 1 day, also with a single 2080 Ti GPU.

\section*{Appendix E: Potential Applications}
\begin{figure}[h]
    \centering
    \includegraphics[width=0.4\textwidth]{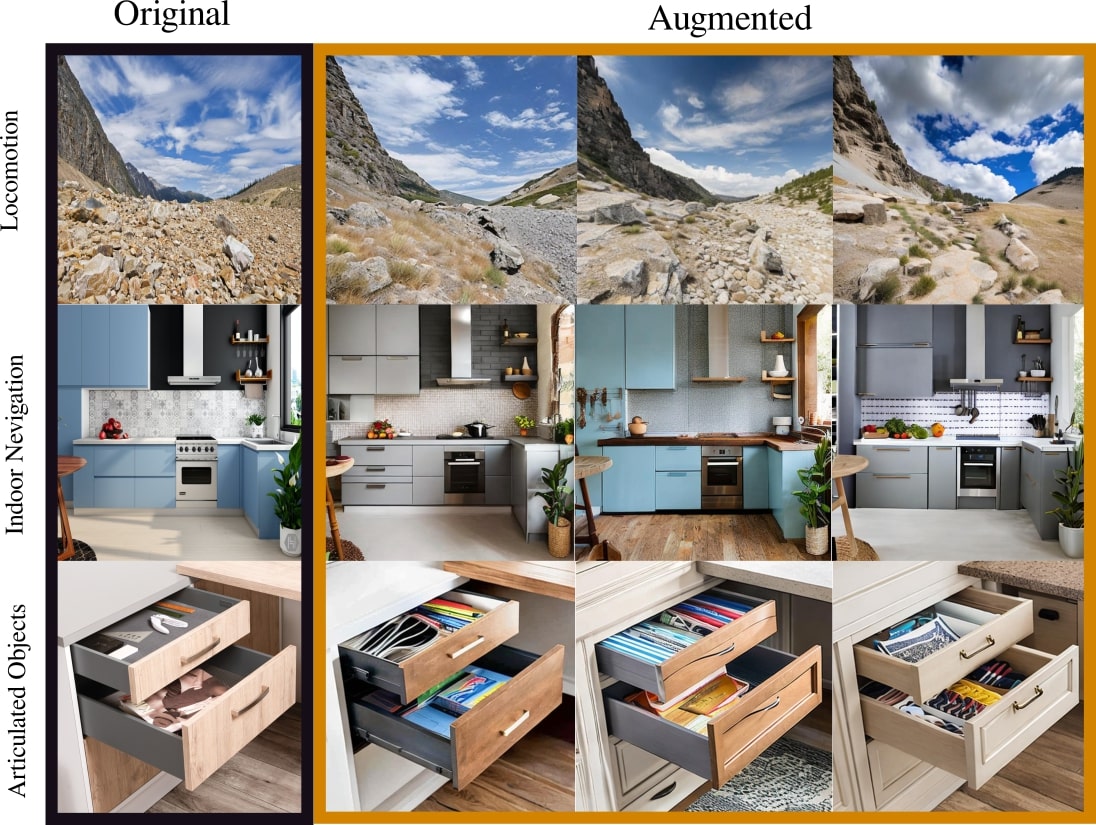}
    \vspace{-0em}\caption{Potential use of our framework for other tasks such as locomotion, indoor navigation, or articulated object manipulation}
    \vspace{-0em}
    \label{fig:other_tasks}
\end{figure}
Our framework is versatile and general and can be potentially applied to other robotics domains such as locomotion, indoor navigation, and articulated object manipulation. Figure \ref{fig:other_tasks} visualizes some examples of how to introduce visual variance to these tasks.
 
\end{document}